\documentclass{article}
\usepackage{amsmath}
\usepackage[nonatbib,preprint]{neurips_2018}

\usepackage[utf8]{inputenc} % allow utf-8 input
\usepackage[T1]{fontenc}    % use 8-bit T1 fonts
\usepackage{hyperref}       % hyperlinks
\usepackage{url}            % simple URL typesetting
\usepackage{booktabs}       % professional-quality tables
\usepackage{amsfonts}       % blackboard math symbols
\usepackage{nicefrac}       % compact symbols for 1/2, etc.
\usepackage{microtype}      % microtypography
\usepackage{todonotes}

\usepackage{latexsym}
\usepackage{amsmath,amsfonts,mathrsfs,amssymb,bm,mathtools,mathabx}
\usepackage{multirow,mdframed}
\usepackage{booktabs}
\usepackage{xspace}
\usepackage[export]{adjustbox}% http://ctan.org/pkg/adjustbox
\usepackage{cleveref}
\usepackage[ruled,linesnumbered,noresetcount,vlined]{algorithm2e}
\usepackage{float}
\floatstyle{ruled}
\newfloat{model}{thp}{lop}
\floatname{model}{Model}

\DeclareMathOperator*{\argmax}{argmax}
\DeclareMathOperator*{\argmin}{argmin}

\newcommand{\cO}{\mathcal{O}}
\newcommand{\cD}{\mathcal{D}}

\newcommand{\cM}{\mathcal{M}}
\newcommand{\cL}{\mathcal{L}}
\newcommand{\cN}{\mathcal{N}}
\newcommand{\cJ}{\mathcal{J}}
\newcommand{\cP}{\mathcal{P}}
\newcommand{\cC}{\mathcal{C}}

\newcommand{\by}{\bm{y}}

\let\oldnl\nl% Store \nl in \oldnl
\newcommand{\nonl}{\renewcommand{\nl}{\let\nl\oldnl}}% Remove line number for one line

\newcommand{\B}{{\text{B}}}
\newcommand{\C}{{\text{C}}}

\newcommand{\CD}{{\text{C}}^{\text{D}}}

\title{Lagrangian Duality for Constrained Deep Learning}

\author{%
  Ferdinando Fioretto \\
  {\small Syracuse University} \\
  \texttt{ffiorett@syr.edu}\\
  \And
  Pascal Van Hentenryck\\
  {\small Georgia Institute of Technology}\\
  \texttt{pvh@isye.gatech.edu}
  \And
  Terrence W.K.~Mak\\
  {\small Georgia Institute of Technology}\\
  \texttt{{wmak@gatech.edu}}
  \AND
  Cuong Tran  \\
  {\small Syracuse University} \\
  \texttt{cutran@syr.edu} \\
  \And 
  Federico Baldo\\
  {\small University of Bologna}\\
  \texttt{\small{federico.baldo2@unibo.it}}
  \And 
  Michele Lombardi\\
  {\small University of Bologna}\\
  \texttt{\small{michele.lombardi2@unibo.it}}
}

\begin{document}
% \nipsfinalcopy is no longer used

\maketitle\sloppy\allowdisplaybreaks
\begin{abstract}
This paper explores the potential of Lagrangian duality for learning applications that feature complex constraints. Such constraints arise in many science and engineering domains, where the task amounts to learning optimization problems which must be solved repeatedly and include hard physical and operational constraints. The paper also considers applications where the learning task must enforce constraints on the predictor itself, either because they are natural properties of the function to learn or because it is desirable from a societal standpoint to impose them.  

This paper demonstrates experimentally that Lagrangian duality brings significant benefits for these applications. In energy domains, the combination of Lagrangian duality and deep learning can be used to obtain state of the art results to predict optimal power flows, in energy systems, and optimal compressor settings, in gas networks. In transprecision computing, Lagrangian duality can complement deep learning to impose monotonicity constraints on the predictor without sacrificing accuracy. Finally, Lagrangian duality can be used to enforce fairness constraints on a predictor and obtain state-of-the-art results when minimizing disparate treatments. 
\end{abstract}

\section{Introduction}

Deep Neural Networks, in conjunction with progress in GPU technology
and the availability of large data sets, have proven enormously successful at a wide array of tasks, including image classification \cite{alexnet}, speech recognition \cite{amodei2016deep}, and natural language processing \cite{collobert2008unified}, to name but a few examples. 
More generally, deep learning has achieved significant success on a variety of regression and classification tasks.
On the other hand, the application of deep learning to aid  computationally challenging constrained optimization problems has been more sparse, but is receiving increasing attention, such as the efforts in jointly training prediction and optimization models \cite{pointer_nets,khalil2017learning,kool2018attention} and incorporating optimization algorithms into differentiable systems \cite{donti2017task,amos2017optnet,wilder2019melding}.

{\em This research originated in an attempt to apply deep learning to fundamentally different application areas: The learning of constrained optimization problems and, in particular, optimization problems with hard physical and engineering constraints.} 
These constrained optimization problems arise in numerous contexts including in energy systems, mobility, resilience, and disaster
management. Indeed, these applications must capture physical laws such as Ohm's law and Kirchhoff's law in electrical power systems, the
Weymouth equation in gas networks, flow constraints in transportation
models, and the Navier-Stoke's equations for shallow water in flood
mitigation. Moreover, they often feature constraints that represent good engineering and operational practice to
protect various devices. For instance, they may include thermal
limits, voltage and pressure bounds, as well as generator and pump
limitations, when the domain is that of energy systems. 
Direct applications of deep learning to these
applications may result in predictions with severe constraint
violations, as shown in Section \ref{sec:results}.

There is thus a need to provide deep learning architectures with capabilities that would allow them to capture constraints directly. Such models can have a transformative impact in many engineering applications by providing high-quality solutions in real-time and be a cornerstone for large planning studies that run
multi-year simulations. 
To this end, this paper proposes a \emph{Lagrangian Dual Framework} (LDF) for Deep Learning that addresses the challenge of enforcing constraints during learning:
Its key idea is to exploit Lagrangian duality, which is widely used to obtain tight bounds in optimization, during the training cycle of a deep learning model. 

{\em Interestingly, the proposed LDF can be applied to two distinct context: 
(1) constrained optimization problems, which are characterized by constraints modeling relations among features of each data sample, and 
(2) problems that require specific properties to hold on the predictor itself, called \emph{constrained predictor problems}}. 
For instance, energy optimization problems are example problems of the first class. These problems impose constraints that are specific to each data sample, such as, flow conservation constraints or thermal limits bounds. 
An example problem of the second class is \emph{transprecision computing}, a technique that achieves energy savings by adjusting the precision of power-hungry algorithms. An important challenge in this area is to predict the error resulting from a loss in accuracy and the error should be monotonically decreasing with increases in accuracy. As a result, the learning task may impose constraints over different samples with their predictions used during training. 
Other applications in dataset-dependent constraint learning may 
impose fairness constraints on the predictor, e.g., a constraint ensuring equal opportunity \cite{NIPS2016_6374} or no disparate impact \cite{zafar2019} in a classifier that enforces a relation among multiple samples of the dataset. 

This paper shows that the proposed LDF provides a versatile tool to address these constrained learning problems, it presents the theoretical foundations of the proposed framework, and demonstrates its practical potential on both constrained optimization and constrained predictors problems. The LDF is evaluated extensively on a variety of real benchmarks in power system optimization and gas compression optimization, that present hard engineering and operational constraints. Additionally, the proposed method is tested on several datasets that enforce non-discriminatory decisions and on a realistic transprecision computing application, that requires constraints to be enforced on the predictors themselves. 
The results present a dramatic improvement in the number of constraint violations reduction, and often result in substantial improvements in the prediction accuracy in energy optimization problems. 

%%%%%%%%%%%%%%%%%%%%%%%%%%%%%%%%%%%%
\section{Preliminaries: Lagrangian Duality}
%%%%%%%%%%%%%%%%%%%%%%%%%%%%%%%%%%%%

Consider the optimization problem
\begin{equation}
\label{eq:problem}
  {\cal O} = \argmin_{y} f(y) \;\;
  \mbox{ subject to } \;\; g_i(y) \leq 0 \;\; (\forall i \in [m]).
\end{equation}
In \emph{Lagrangian relaxation}, some or all the problem constraints
are relaxed into the objective function using \emph{Lagrangian
  multipliers} to capture the penalty induced by violating them. 
  When all the constraints are relaxed, the \emph{Lagrangian
  function} becomes
\begin{equation}
\label{eq:lagrangian_function}
f_{\lambda}(y) = f(y) + \sum_{i=1}^m \lambda_i g_i(y)
\end{equation}
where the terms $\lambda_i \geq 0$ describe the Lagrangian multipliers, and $\lambda = (\lambda_1, \ldots, \lambda_m)$ denotes the vector of all multipliers associated to the problem constraints.
Note that, in this formulation, $g(y)$ can be positive or negative. 
An alternative formulation, used in augmented Lagrangian methods 
\cite{Hestenes:69} and constraint programming \cite{Fontaine:14}, 
uses the following Lagrangian function 
\begin{equation}
\label{eq:lagrangian_function2}
f_{\lambda}(y) = f(y) + \sum_{i=1}^m \lambda_i \max(0, g_i(y))
\end{equation}
where the expressions $\max(0, g_i(y))$ capture a quantification of the constraint violations. This paper abstracts the constraints 
formulations in \eqref{eq:lagrangian_function} and \eqref{eq:lagrangian_function2}
by using a function $\nu(\cdot)$ that returns either the constraint
satisfiability or the violation degree of a constraint.  

When using a Lagrangian function, the optimization problem becomes
\begin{equation}
\label{eq:LR}
LR_{\lambda} = \argmin_y f_{\lambda}(y)
\end{equation}
and it satisfies $f(LR_{\lambda}) \leq f({\cal O})$. That is, 
the Lagrangian function is a lower bound for the original function. 
Finally, to obtain the strongest Lagrangian relaxation of ${\cal O}$,
the \emph{Lagrangian dual} can be used to find the best Lagrangian
multipliers, i.e.,
\begin{equation}
\label{eq:LD}
LD = \argmax_{\lambda \geq 0} f(LR_{\lambda}).
\end{equation}
For various classes of problems, the Lagrangian dual is a strong
approximation of ${\cal O}$. Moreover, its optimal solutions can often
be translated into high-quality feasible solutions by a
post-processing step, i.e., using a \emph{proximal operator} that 
minimizes the changes to the Lagrangian dual solution while projecting  it into the problem feasible region \cite{parikh2014proximal}.

%%%%%%%%%%%%%%%%%%%%%%%%%%%%%%%%%%%%
\section{Learning Constrained Optimization Problems}
%%%%%%%%%%%%%%%%%%%%%%%%%%%%%%%%%%%%

This section describes how to use the {Lagrangian dual framework} for approximating constrained optimization problems in which constraints model relations among features of each data sample. Importantly, in the associated learning task, each data sample represents a different instantiation of a constrained optimization problem. The section first reviews two fundamental applications that serve as motivation.

%%%%%%%%%%%%%%%%%%%%%%%%%%%%%%%%%%%%
\subsection{Motivating Applications}
%%%%%%%%%%%%%%%%%%%%%%%%%%%%%%%%%%%%

Several energy systems require solving challenging (non-convex,
non-linear) optimization problems in order to derive the best system
operational controls to serve the energy demands of the
customers. Power grid and gas pipeline systems are two examples of 
such applications. 
While these problems can be solved using effective
optimization solvers, their resolution relies on \emph{accurate}
predictions of the energy demands. The increasing penetration of
renewable energy sources, including those behind the meter (e.g., 
solar panels on roofs), has rendered accurate predictions more 
challenging. In turn, predictions need to be performed at minute time 
scales to ensure sufficient accuracy. 
Thus, finding optimal solutions for these underlying optimization problems
in these reduced time scales becomes computationally challenging, 
opening opportunities for machine-learning approaches. The next 
paragraphs review two energy applications that motivate the proposed 
framework. An extended description of these models is provided in the supplemental material. 

%%%%%%%%%%%%%%%%%%%%%%%%%%%%%%%%%%%%
\paragraph{Optimal Power Flow} 
%%%%%%%%%%%%%%%%%%%%%%%%%%%%%%%%%%%%
The \emph{Optimal Power Flow} (OPF)
problem determines the best generator dispatch $(y = S^g)$ of minimal cost $({\cal O} = \min_{S^g} \mbox{cost}(S^g))$
that meets the demands $(d = S^d)$ while satisfying the physical and engineering constraints ($g(y)$) of the power system \cite{OPF}, where $S^g$ and $S^d$ denote the vectors (in the complex domain) of generator dispatches and power demands. 
Typical constraints include the non-linear non-convex AC power flow equations, Kirchhoff's current laws,
voltage bounds, and thermal limits.
The OPF problem is a fundamental building bock of many
applications, including security-constrained OPFs \cite{monticelli:87}), 
optimal transmission switching \cite{OTS}, capacitor placement \cite{baran:89}, 
and expansion planning \cite{verma:16} %and security-constrained unit commitment \cite{wang:08}.
which are of fundamental importance for ensuring a reliable and efficient 
behavior of the energy system.

%%%%%%%%%%%%%%%%%%%%%%%%%%%%%%%%%%%%
\paragraph{Optimal Gas Compressor Optimization}
%%%%%%%%%%%%%%%%%%%%%%%%%%%%%%%%%%%%
The Optimal Gas Compressor Optimization (OGC) problem aims at
determining the best compression controls ($y=R$) 
with minimum compression costs $({\cal O} = \min_R \text{cost}(R))$
to meet gas demands ($d=q^d$) while
satisfying the physical and operational limits ($g(y)$) of the natural gas pipeline systems~\cite{herty2010new}. Therein, $R$ and $q^d$ are compressors control values and gas demands. 
Typical constraints include: the non-linear gas flow equations describing pressure losses,
the flow balance equations, the non-linear non-convex compressor objective ${\cal O}$, and the pressure bounds. 
Similar to the OPF problem, the OGC is a non-linear non-convex optimization problem with physical and engineering constraints and a fundamental building block for many gas systems.

The next section describes how to approximate OPFs and OGCs, by viewing them as parametric optimization problems, using the proposed Lagrangian dual framework.

%%%%%%%%%%%%%%%%%%%%%%%%%%%%%%%%
\subsection{The Learning Task}
\label{sec:learning_task}
%%%%%%%%%%%%%%%%%%%%%%%%%%%%%%%%%%%%
The learning task estimates a parametric version of problem \eqref{eq:problem}, defined as
\begin{equation}
\label{eq:par_problem}
  {\cal O}(d) = \argmin_{y} f(y,d) \;\;
  \mbox{ subject to } \;\; g_i(y,d) \leq 0
  \;\; (\forall i \in [m])
\end{equation}
with a set of samples $D = \left\{ (d_l,y_l={\cal O}(d_l))\right\}_{l=1}^n$. 
More precisely, given a parametric model ${\cal M}[w]$ with weights $w$ and a loss function ${\cal L}$, the learning task must solve the following optimization problem
\begin{subequations}
\label{eq:learning_problem}
\begin{flalign}
  w^* = & \argmin_w  \sum_{l=1}^{n} {\cal L}({\cal M}[w](d_l),y_l) \\
        & \mbox{ subject to } \;
        g_i({\cal M}[w](d_l),d_l) \leq 0 \;\; (\forall i \in [m], l \in [n])
\end{flalign}
\end{subequations}
to obtain the approximation $\widehat{{\cal O}} = {\cal M}[w^*]$ of
${\cal O}$. 

The main difficulty lies in the constraints $g_i(y,d) \leq
0$, which can represent physical and operational limits, as
mentioned in the motivating applications. Observe that
the model weights must be chosen so that the constraints are satisfied for all samples, which makes the learning particularly 
challenging. A naive approach to the learning task is thus likely to result in
predictors that significantly violate these constraints, {as
demonstrated in Section \ref{sec:results}}, producing a model that
would not be useful in practice.

%%%%%%%%%%%%%%%%%%%%%%%%%%%%%%%%%%%%
\subsection{Lagrangian Dual Framework for Constrained Optimization Problems}
\label{sec:lagrangian_duality}
%%%%%%%%%%%%%%%%%%%%%%%%%%%%%%%%%%%%
To learn constrained optimization problems, the paper proposes a
\emph{Lagrangian dual framework} (LDF) to the learning task. The framework relies on the notion of \emph{Augmented Lagrangian}~\cite{Hestenes:69} used for solving constrained optimization problems \cite{Fontaine:14}. 

In more details, LDF exploits a Lagrangian dual approach in the learning task to approximate the minimizer ${\cal O}$. Given multipliers $\lambda = (\lambda_1, \ldots, \lambda_m)$, consider the Lagrangian loss function
\begin{equation*}
\label{eq:Lagr_loss}
  {\cal L}_{\lambda}(\hat{y}_l,y_l,d_l) = {\cal L}(\hat{y}_l,y_l) 
  + \sum_{i=1}^m \lambda_i \, \nu\left(g_i (\hat{y}_l, d_l)\right),%, \leq 0).
\end{equation*}
where $\hat{y}_l = {\cal M}[w](d_l)$ represents the model prediction. 
For multipliers $\lambda$, solving the optimization problem
\begin{equation}
\label{eq:lagr_opt}
w^*(\lambda) = \argmin_w \sum_{l=1}^{n} {\cal L}_{\lambda}({\cal M}[w](d_l),y_l,d_l)
\end{equation}
produces an approximation $\widehat{{\cal O}}_{\lambda} = {\cal
  M}[w^*(\lambda)]$ of ${\cal O}$. The Lagrangian dual computes the
optimal multipliers, i.e.,
\begin{equation}
\label{eq:opt_multipliers}
\lambda^* = \argmax_{\lambda} \min_w \sum_{l=1}^{n} {\cal L}_{\lambda}({\cal M}[w](d_l),y_l,d_l)
\end{equation}
to obtain $\widehat{{\cal O}}^* = {\cal M}[w^*(\lambda^*)]$, i.e.,
the strongest Lagrangian relaxation of ${\cal O}$. 

Learning $\widehat{{\cal O}}^*$ relies on an iterative scheme that
interleaves the learning of a number of Lagrangian relaxations (for
various multipliers) with a subgradient method to learn the best
multipliers. 
The LDF, described in Equations \eqref{eq:lagr_opt} and \eqref{eq:opt_multipliers}, is summarized in Algorithm \ref{alg:learning}. 
Given the input dataset $D$, the optimizer step size $\alpha > 0$, and a Lagrangian step size $s_k$, the Lagrangian multipliers are initialized in line \ref{line:1}. The training is performed for a fixed number of epochs, and each epoch $k$ optimizes the model weights $w$ of the optimizer ${\cal M}[w, \lambda^k]$ using the Lagrangian multipliers $\lambda^k$ associated with current epoch (lines \ref{line:3}--\ref{line:5}). 
Finally, after each epoch, the Lagrangian multipliers are updated according to a \emph{dual ascent} rule \cite{boyd2011distributed} (line \ref{line:6}).

\begin{algorithm}[!t]
{%\small
  \caption{LDF for Constrained Optimization Problems}
  \label{alg:learning}
  \setcounter{AlgoLine}{0}
  \SetKwInOut{Input}{input}

  \Input{$D=(d_l, y_l)_{l=1}^n:$ Training data; \\
       $\alpha, s=(s_0, s_1,\ldots):$ Optimizer and Lagrangian step sizes.\!\!\!\!\!\!\!\!\!\!}
  \label{line:1}
  $\lambda_i^0 \gets 0 \;\; \forall i \in [m]$\\
  \For{epoch $k = 0, 1, \ldots$} { 
  \label{line:2}
      \ForEach{$(y_l, d_l) \in D$} { %\!\gets\! \mbox{minibatch}(D)$ of size $b$}{
      \label{line:3}
      $\hat{y}_l \gets {\cal M}[w, \lambda^k](d_l)$\\
      \label{line:4}
      $w \gets w - \alpha \nabla_{w} 
        {\cal L}_{\lambda^k}(\hat{y}_l, y_l, d_l) $
      \label{line:5}
    }
    $\lambda^{k+1}_i \gets \lambda^k_i + s_k \sum_{l=1}^n 
    \nu_i \left( g_i(\hat{y}_l, d_l) \right) \;\;
    \forall i \in [m]$
    \label{line:6}
  }
}
\end{algorithm}

%%%%%%%%%%%%%%%%%%%%%%%%%%%%%%%%%%%%
\section{Learning Constrained Predictors}
\label{sec:constrained_learning}
%%%%%%%%%%%%%%%%%%%%%%%%%%%%%%%%%%%%

This section describes how to use the Lagrangian dual framework for problems in which constraints are not sample-independent, but enforcing global properties between different samples in the dataset and the predictor outputs. It starts with two motivating applications.

%%%%%%%%%%%%%%%%%%%%%%%%%%%%%%%%%%%%
\subsection{Motivating Applications}
\label{sec:const_learning}
%%%%%%%%%%%%%%%%%%%%%%%%%%%%%%%%%%%%

Several applications require to enforce constraints on the
learning process itself to attain desirable properties of the
predictor. These constraints impose conditions on subsets of the
samples that must be satisfied. For instance, assume that there is a partial order $\preceq$
on the optimization inputs and the following property holds:
\[
d_1 \preceq d_2 \Rightarrow f({\cal O}(d_1),d_1) \leq f({\cal O}(d_2),d_2).
\]
The predictor should ideally satisfy these constraints as well:
\[
d_1 \preceq d_2 \Rightarrow f(\widehat{{\cal O}}(d_1),d_1) \leq f(\widehat{{\cal O}}(d_2),d_2).
\]

%%%%%%%%%%%%%%%%%%%%%%%%%%%%%%%%%%%%
\paragraph{Transprecision computing}
%%%%%%%%%%%%%%%%%%%%%%%%%%%%%%%%%%%%
Transprecision computing is the idea of reducing energy consumption by
reducing the precision (a.k.a.~number of bits) of the variables
involved in a computation \cite{malossi2018transprecision}.  It is
especially important in low-power embedded platforms, which arise in
many contexts such as smart wearable and autonomous vechicles.
Increasing precision typically reduces the error of the target
algorithm. However, it also increases the energy consumption, which is
a function of the maximal number of used bits.  The objective is to
design a \emph{configuration} $d_l$, i.e., a mapping from input
computation to the precision for the variables involved in the
computation.  The sought configuration should balance \emph{precision}
and \emph{energy consumption}, given a bound to the error produced by
the loss in precision when the highest precision configuration is
adopted.

However, given a configuration, computing the corresponding error can
be very time-consuming and the task considered in this paper seeks to
learn a mapping between configurations and error. This learning task
is non-trivial, since the solution space precision-error is non-smooth
and non-linear \cite{malossi2018transprecision}.  The samples $(d_l,
y_l)$ in the dataset represent, respectively, a configuration $d_l$
and its associated error $y_l$ obtained by running the configuration
$d_l$ for a given computation.  The problem $\cO(d_l)$ specifies the
error obtained when using configuration $d_l$.

Importantly, transcomputing expects a \emph{monotonic} behavior:
Higher precision configurations should generate more accurate results
(i.e., a smaller error).  Therefore, the structure of the problem imposes the
learning task to require a dominance relation $\preceq$ between
instances of the dataset. More precisely, $d_2 \preceq d_2$ holds
if 
\[
\forall i \in [N]: \ x_{1_i} \leq x_{2_i}
\]
where $N$ is the number of variables involved in the computation and
$x_{1_i}$, $x_{2_i}$ are the precision values for the variables in
$d_1$ and $d_2$ respectively.  %For instance, if $d_1 = [43, 23, 17, 50]$ and $d_2 = [27, 22, 10, 4]$, then $d_2 \preceq d_2$.  

\paragraph{Fair Classifier}
The second motivating application considers the task of building 
a classifier that satisfies \emph{disparate impact} \cite{zafar2019} 
with respect to a protected attribute $d^s$ and outcome $y$. A binary classifier 
does not suffer from disparate impact if
\begin{equation}
\label{eq:eq_opp}
    \Pr(\hat{y} = 1 | d^s = 0) = \Pr(\hat{y} = 1 | d^s = 1).
\end{equation}
For outcome $y=1$, the constraint above requires the predictor
$\hat{y}$ to have equal \emph{predicted positive rates} across the 
different sensitive classes: $d^s=0$ and $d^s=1$, in the binary 
task example above. For $y=0$, the constraint enforces equal 
\emph{predicted negative rates}. Disparate impact constraints 
the predicted positive (or negative) rates to be similar across 
all sensitive attributes. 
To construct an estimator that minimizes the disparate impact, the paper considers $|\cD_s|=2$ estimators $\cM_{0}$ and $\cM_{1}$, each associated with a dataset partition $D_{|s_i} = \{(d_l, y_l) | d_l^s = s_i\}$ that marginalizes for a particular (combination of) value(s) of the protected feature(s), in addition to the classical estimator $\cM$ that is trained over the entire dataset $D$. 
Thus, the learning process is defined by the following objective:
\begin{subequations}
\begin{align}
  \min_{w, w_0, w_1} & 
           \cL\left(\mathcal{M}[w](D)\right) 
  + \sum_{i=0}^1 \cL\left(\mathcal{M}_i[w_i](D_{|s_i})\right)
  \label{farines_1}
  \\
  \text{such that~}& 
  \frac{\left|\sum_{x_i \in D_{s_0}} I(\hat{y}_i =1) \right|}{\left|D_{s_0}\right|}
  - 
  \frac{\left|\sum_{x_i \in D_{s_1}} I(\hat{y}_i =1) \right|}{\left|D_{s_1}\right|},
    \label{farines_2}
\end{align}
\end{subequations}
where $I$ is the indicator function. It enforces a constraint on the output of the classifiers $\cM_0$, trained on data $D_{s_0}$ to be equivalent to that of the output of the classifier $\cM_1$,  trained on the dataset $D_{s_1}$, when their predicted outcome is positive. 

The next section will specify how to encode such type of constraints as well as how to express and enforce dominance relations in the proposed constrained learning framework.

%%%%%%%%%%%%%%%%%%%%%%%%%%%%%%%%%%%%
\subsection{The Learning Task}
%%%%%%%%%%%%%%%%%%%%%%%%%%%%%%%%%%%%

Consider a set ${\cal S} = \{ S_1, \ldots, S_q \}$ where $S_i$ is a
subset of the inputs that must satisfy the associated constraint
\[
h_i(\{{\cal O}(d_l)\}_{l \in S_i}, \bm{d}_{S_i}),
\]
where $\bm{d}_{S_i} = \{d_l\}_{l \in S_i}$, and denote $\bm{y}_{S_i} = \{y_l\}_{l \in S_i}$.

In this context, the learning task is defined by the following optimization problem
\begin{subequations}
\label{eq:constr_learn}
\begin{flalign}
  \argmin_w \; & \sum_{l=1}^n {\cal L}({\cal M}[w](d_l),y_l) \\
  \mbox{ subject to } \; 
  & g_i\left({\cal M}[w](d_l),d_l\right) \leq 0 \;\; (\forall i \in [m], l \in [n]) \\
  & h_i\left(\{{\cal M}[w](d_l)\}_{l \in S_i}, \bm{d}_{S_i}\right) \;\; (\forall i \in [q]).
\end{flalign}
\end{subequations}

%%%%%%%%%%%%%%%%%%%%%%%%%%%%%%%%%%%%
\subsection{Lagrangian Dual Framework for Constrained Predictors}
%%%%%%%%%%%%%%%%%%%%%%%%%%%%%%%%%%%%

To approximate Problem \eqref{eq:constr_learn}, the learning task considers Lagrangian loss functions, for subset of the inputs $S_i \in {\cal S}$, of the form
\begin{equation}
  {\cal L}_{\mu,\lambda}(\tilde{\bm{y}}_{S_i}, \bm{y}_{S_i}, \bm{d}_{S_i}) = 
  \sum_{l \in S_i} {\cal L}_{\lambda}(\tilde{y}_l, y_l, d_l) + 
  \sum_{i=1}^q \mu_i \ \nu 
  \left( h_i(\tilde{\bm{y}}_{S_i}, \bm{d}_i)\right),
\end{equation}
where $\tilde{y}_l = \cM[w](d_l)$ 
and $\tilde{\bm{y}}_{S_i} = \{ {\cal M}[w](d_l)_{l \in S_i} \}$. 
It learns approximations of the Lagrangian relaxations $\widehat{{\cal O}}_{\lambda,\mu}$ of the form
\begin{equation}
\label{eq:CL_w}
  w^*(\mu,\lambda) = \argmin_w 
    \sum_{i=1}^{q} {\cal L}_{\mu,\lambda}
    (\{{\cal M}[w](d_l)\}_{l \in S_i}, \bm{y}_{S_i}, \bm{d}_{S_i}),
\end{equation}
as well as the Lagrangian duals of Equation \eqref{eq:CL_w} of the form 
\begin{equation}
\label{eq:CL_lambda}
  \lambda^*(\mu) = \argmax_{\lambda} \min_w 
  \sum_{i=1}^{q} {\cal L}_{\mu,\lambda}
  (\{ {\cal M}[w](d_l)\}_{l\in S_i} ,\bm{y}_{S_i}, \bm{d}_{S_i}),
\end{equation}
and, finally, the Lagrangian dual of the Lagrangian duals (Equation \eqref{eq:CL_lambda}) as
\begin{equation}
\label{eq:CL_mu}
  \mu^* = \argmax_{\mu} \max_{\lambda} \min_w 
  \sum_{i=1}^{q} {\cal L}_{\mu,\lambda}
  (\{ {\cal M}[w](d_l)\}_{l\in S_i} ,\bm{y}_{S_i}, \bm{d}_{S_i}) 
\end{equation}
to obtain the best estimator $\widehat{{\cal O}}^* = {\cal M}[w^*]$, where
\[
  w^* = \argmin_w \sum_{i=1}^{q} 
  {\cal L}_{\mu^*,\lambda^*(\mu^*)}
  (\{ {\cal M}[w](d_l)\}_{l\in S_i} ,\bm{y}_{S_i}, \bm{d}_{S_i}).
\]

The Lagrangian dual framework for constrained predictors, described in Equations \eqref{eq:CL_w}--\eqref{eq:CL_mu}, is summarized in Algorithm \ref{alg:learning2}.
The learning algorithm interleaves the learning of the Lagrangian duals with the subgradient optimization of the multipliers $\mu$. 
Given the input dataset $D$, a set ${\cal S}$ of subsets of inputs, the optimizer step size $\alpha > 0$, and Lagrangian step sizes $s_k$, and $t_k$, the Lagrangian multipliers are initialized in lines \ref{line:1b} and \ref{line:1a}. 
The training is performed for a fixed number of epochs, and each epoch $k$ optimizes the model weights $w$ of the optimizer ${\cal M}$ using the Lagrangian multipliers $\lambda^k$ and $\mu^k$ associated with current epoch $k$, denoted ${\cal M}[w, \lambda^k, \mu^k]$ in the algorithm (lines \ref{line:3a}--\ref{line:5a}). 
Similarly to Algorithm \ref{alg:learning}, the Lagrangian multipliers $\lambda_i$ for the dual variables are updated after each epoch, on line \ref{line:6a}). Finally, the algorithm updates the multipliers $\mu_i$ associated to the Lagrangian duals of the Lagrangian duals (line \ref{line:7a}).

\begin{algorithm}[!t]
{%\small
  \caption{LDF for Constrained Predictor Problems}
  \label{alg:learning2}
  \setcounter{AlgoLine}{0}
  \SetKwInOut{Input}{input}

  \Input{$D=(d_l, y_l)_{l=1}^n, {\cal S} = \{S_1, \ldots, S_n\}$ Training data and data partitions;\\
     $\alpha, s=(s_0, s_1,\ldots), t=(t_0, t_1,\ldots):$ Optimizer and Lagrangian step sizes.\!\!\!\!\!\!\!\!\!\!}
  \label{line:1a}
  $\lambda_i^0 \gets 0 \;\; \forall i \in [m]$\\
  \label{line:1b}
  $\mu_i^0 \gets 0 \;\; \forall i \in [q]$\\
  \label{line:2a}
  \For{epoch $k = 0, 1, \ldots$} {
      \ForEach{$S_i \in {\cal S}$} 
      { 
        % \ForEach{$(y_l, d_l) \in D$} { 
        \label{line:3a}
        $\hat{\bm{y}}_{S_i} \gets 
        \{ {\cal M}[w, \lambda^k, \mu^k] (d_l) \}_{l \in S_i}$\\
        \label{line:4a}
        $w \gets w - \alpha \nabla_{w} 
          {\cal L}_{\lambda^k, \mu^k}(\hat{\bm{y}}_{S_i}, \bm{y}_{S_i}, \bm{d}_{S_i}) $
        \label{line:5a}
      }
    $\lambda^{k+1}_i \gets \lambda^k_i + s_k \sum_{l=1}^n 
    \nu_i \left( g_i(\hat{y}_l, d_l) \right) \;\; \forall i \in [m]$\\    
    \label{line:6a}
    $\mu^{k+1}_i \gets \mu^k_i + t_k  
    \nu_i \left( h(\hat{\bm{y}}_{S_i},\bm{d}_{S_i}) \right) \;\; 
    \forall i \in [q]$\\    
    \label{line:7a}
  }
}
\end{algorithm}

%%%%%%%%%%%%%%%%%%%%%%%%%%%%%%%%%%%%
\section{Experiments}
\label{sec:results}
%%%%%%%%%%%%%%%%%%%%%%%%%%%%%%%%%%%%

This section evaluates the proposed LDF on constrained optimization problems for energy and gas networks and on constrained learning problems--that enforce constraints on the predictors--for applications in transprecision computing and fairness.

%%%%%%%%%%%%%%%%%%%%%%%%%%%%%%%%%%%%
\subsection{Constrained Optimization Problems}
%%%%%%%%%%%%%%%%%%%%%%%%%%%%%%%%%%%%

\smallskip\noindent\textbf{Data set Generation}
The experiments examine the proposed models on a variety of power
networks from the NESTA library \cite{Coffrin14Nesta}
and natural gas benchmarks from~\cite{mak19dynamic} and GasLib~\cite{pfetsch2015validation}.
% presentation simplicity, the analysis focuses primarily on the IEEE
% 30, 118, and 300-bus networks. However, the results are consistent
% across the entire benchmark set (see \cite{Fioretto:dnnopf}).  
The ground truth data are constructed as follows: For each power and gas network, different benchmarks are generated by altering the amount of nominal demands $d \!=\! S^d$ (for power networks) and $d \!=\! q^d$ (for gas networks) within a $\pm 20\%$ range. 
The resulting 4000 demand vectors are used to generate solutions to the OPF and OGC problems. Increasing loads causes heavily congestions to the system, rendering the computation of optimal solutions challenging. A network value, that constitutes a dataset entry
 $(d_l, y_l=\cO(d))$, is a feasible solution obtained by solving
 the AC-OPF problem \cite{OPF}, for electricity networks, or the OGC problem, for gas networks \cite{herty2010new}. The experiments use a $80/20$ train-test split and results are reported on the test set.

\smallskip\noindent\textbf{Learning Models}
The experiments use a baseline ReLU network ${\cal M}$, with 5 layers which minimizes the Mean Squared Error (MSE) loss ${\cal L}$ to predict to active power $\hat{p}$, voltage magnitude $\hat{v}$, and voltage angle $\hat{\theta}$, for energy networks, and 
compression ratios $\hat{R}$, pressure $\hat{p}$, and gas flows $\hat{q}$, for gas networks. 

This baseline model is compared with a model ${\cal M}_\C$ that exploits the problem constraints and minimizes the loss: ${\cal L} + \lambda \nu(\cdot)$, with multiplier values $\lambda$ fixed to $\bm{1}$.  
Finally, ${\cal M}_\CD$ extends model ${\cal M}_\C$ by learning the Lagrangian multipliers using the LDF introduced in Section \ref{sec:lagrangian_duality}.
The constrained learning model for power systems also
exploits the hot-start techniques used
in~\cite{ferdin2020predicting}, with states differing by at most
$1\%$. Experiments using larger percentages (up to 3\%) showed  similar trends. 
The training uses the Adam optimizer with learning rate ($\alpha=10^{-3}$) and was performed for $80$ epochs using batch sizes $b=64$. Finally, the Lagrangian step size $\rho$ is set to $10^{-4}$. 
Extensive additional information about the network structure,  the optimization model (OPF and OGC), the learning loss functions and constraints, as well as additional experimental analysis is provided in the appendix.

%%%%%%%%%%%%%%%%%%%%%%%%%%%%%%%%%%%%%%%
\smallskip\noindent\textbf{Prediction Errors}
%%%%%%%%%%%%%%%%%%%%%%%%%%%%%%%%%%%%%%%
%We first analyze the prediction error of the learning models.
Table \ref{tbl:prediction_errors} and \ref{tbl:mean_prediction_errors_gas} 
report the average L$_1$-distance and the \emph{prediction errors} 
between a subset of predicted variables $y$ (marked with $\hat{y}$) 
on both the power and gas benchmarks and their original ground-truth
quantities.  
The error for $y$ is reported in percentage as 
$100 \frac{\|\hat{y} - y\|_1}{\|y\|_1}$ and the gain (in parenthesis) reports the ratio between the error obtained by the baseline model accuracy and the constrained models. 

\begin{table}[!t]
  \centering
  \resizebox{0.65\linewidth}{!}
  {
  \begin{tabular}{l@{\hspace{6pt}} r@{\hspace{2pt}} | @{\hspace{4pt}}
                  r@{\hspace{10pt}} r@{\hspace{10pt}} 
                  l@{\hspace{10pt}} r@{\hspace{10pt}} 
                  l}
  \toprule
  \multirow{2}{*}{\textbf{Test Case}} & 
  \multirow{2}{*}{\textbf{Type}} & 
  ${\cal M}$ &  
  \multicolumn{2}{c}{${\cal M}_\C$} & 
  \multicolumn{2}{c}{${\cal M}_\CD$} \\
  \cmidrule(r){3-3}
  \cmidrule(r){4-5}
  \cmidrule(r){6-7}
  & & err (\%) & err (\%) & gain & err (\%) & gain \\
  %$\hat{{p}}^g$ & $\hat{{v}}$& $\hat{{\theta}}$& $\tilde{{p}}^f$  \\
  % \cmidrule(lr){2-6} 
  \midrule
    \multirow{4}{*}{\textbf{30\_ieee}}
     &   $\hat{p}$      &3.3465 &0.3052  & (10.96) &\textbf{0.0055} & (608.4) \\ 
     &   $\hat{v}$      &14.699 &0.3130  & (46.96) &\textbf{0.0070} & (2099) \\   
     &   $\hat{\theta}$ &4.3130 &0.0580  & (74.36) &\textbf{0.0041} & (1052) \\
     &   $\tilde{p}^f$  &27.213 &0.2030  & (134.1) &\textbf{0.0620} & (438.9) \\
     \hline
     \multirow{4}{*}{\textbf{118\_ieee}}
     &   $\hat{p}$      &0.2150 &0.0380  & (5.658) &\textbf{0.0340} & (6.323) \\
     &   $\hat{v}$      &7.1520 &0.1170  & (61.12) &\textbf{0.0290} & (246.6) \\
     &   $\hat{\theta}$ &4.2600 &1.2750  & (3.341) &\textbf{0.2070} & (20.58) \\
     &   $\tilde{p}^f$  &38.863 &0.6640  & (58.53) &\textbf{0.4550} & (85.41) \\
      \hline
    \multirow{4}{*}{\textbf{300\_ieee}} 
    &    $\hat{p}$      &0.0838 &0.0174  & (4.816) &\textbf{0.0126} & (6.651)\\
    &    $\hat{v}$      &28.025 &3.1130  & (9.002) &\textbf{0.0610} & (459.4)\\
    &    $\hat{\theta}$ &12.137 &7.2330  & (1.678) &\textbf{2.5670} & (4.728)\\
    &    $\tilde{p}^f$  &125.47 &26.905  & (4.663) &\textbf{1.1360} & (110.4)\\
   \bottomrule
  \end{tabular}
  }
  \caption{table}{Mean Prediction Errors (\%) and accuracy gain (\%) on OPF Benchmarks.}
  \label{tbl:prediction_errors} 
\end{table}

For the power networks, the models focus on predicting the active generation dispatches $\hat{{p}}^g = Re(S^g)$, voltage magnitudes $\hat{{v}}$, voltage angles $\hat{{\theta}}$, and the active transmission line (including transformers) flows
$\tilde{{p}}^f$. Power flows $\tilde{{p}}^f$ are not directly predicted but computed from the predicted quantities through the Ohm's laws (See section \ref{sec:opf_details}). 
For the gas networks, the models focus on predicting compression ratios $\hat{{R}}$, pressure values $\hat{{p}}$, and gas flows $\hat{{q}}$. 
%Let $y$ be a quantity to be measured.
The best results are highlighted in bold.  

A clear trend appears: The prediction errors decrease with the
increase in model complexity. In particular, model ${\cal M}_\C$, 
which exploits the problem constraints, predicts voltage quantities 
and power flows that are up to two order of magnitude more precise 
than those predicted by ${\cal M}$, for OPF problems. 
The prediction errors on OGC benchmarks, instead remain of the same 
order of magnitude as those obtained by the baseline model $\cM$, 
albeit the accuracy of the prediction increases consistently when 
adopting the constrained models. 
This can be explained by the fact that the gas networks behave largely 
monotonically in compressor costs for varying loads. 
{\em Finally, the LDF that finds the best multipliers (${\cal M}_\CD$) consistently improves the baseline model 
on OGC benchmarks, and further improves ${\cal M}_\C$ predictions by 
an additional order of magnitude, for OPF problems.\footnote{
The accuracy gains appear more pronounced on OPF problems since 
the baseline model $\cM$ produces already extremely accurate results 
for OGC benchmarks.}}
% The only exception is the 40-pipe gas benchmark, where ${\cal M}$
% gives already extremely accurate results.  

\begin{table}[!t]
\centering
  \resizebox{0.65\linewidth}{!}
  {
  \begin{tabular}{l@{\hspace{6pt}} r@{\hspace{2pt}} | @{\hspace{4pt}}
                  r@{\hspace{10pt}} r@{\hspace{10pt}} 
                  l@{\hspace{10pt}} r@{\hspace{10pt}} 
                  l}
  \toprule
  \multirow{2}{*}{\textbf{Test Case}} & 
  \multirow{2}{*}{\textbf{Type}} & 
  ${\cal M}$ &  
  \multicolumn{2}{c}{${\cal M}_\C$} & 
  \multicolumn{2}{c}{${\cal M}_\CD$} \\
  \cmidrule(r){3-3}
  \cmidrule(r){4-5}
  \cmidrule(r){6-7}
  & & err (\%) & err (\%) & gain & err (\%) & gain \\
  \midrule
  \multirow{4}{*}{\textbf{24-pipe}}
     &   $\hat{R}$ &0.0052     &0.0079 & (0.658) &\bf{0.0025} & (2.080) \\ 
     &   $\hat{p}$ &\bf{0.0057}&0.0068 & (0.838) &\bf{0.0057} & (1.000) \\
     &   $\hat{q}$ &0.0029     &0.0592 & (0.049) &\bf{0.0007} & (4.142) \\
    %  &   ${\cal M}_\CS$ &0.0017&\bf{0.0001}&0.0023 \\
    %  &   ${\cal M}_\CLD$  &\bf{0.0015}&\bf{0.0001}&\bf{0.0019} \\
  \hline
  \multirow{4}{*}{\textbf{40-pipe}}
     &   $\hat{R}$ & 0.0009 & 0.0103 & (0.087) & \bf{0.0006} &(1.833) \\ 
     &   $\hat{p}$ & 0.0011 & 0.0025 & (0.240) & \bf{0.0006} &(1.500) \\
     &   $\hat{q}$ & 0.0006 & 0.0329 & (0.033) & \bf{0.0004} &(1.500) \\
    %  &   ${\cal M}_\CS$ & 0.0007 & \bf{0.0002} & 0.0009\\
    %  &   ${\cal M}_\CLD$  &0.0007 & \bf{0.0002} &0.0012 \\ 
   \hline
       \multirow{4}{*}{\textbf{135-pipe}}
     &   $\hat{R}$ &0.0206 &     0.0317     & (0.650) & \bf{0.0199} & (1.307) \\ 
     &   $\hat{p}$ &0.0260 & \bf{0.0209}    & (1.067) &     0.0225  & (0.916) \\
     &   $\hat{q}$ &0.0223 &     0.0572     & (0.455) & \bf{0.0222} & (1.005) \\
    %  &   ${\cal M}_\CS$ &0.0019 &0.0028 &0.0013 \\
    %  &   ${\cal M}_\CLD$  & \bf{0.0005}& \bf{0.0002}& \bf{0.0006}  \\
   \bottomrule
  \end{tabular}
  }
  \caption{table}{Mean Prediction Errors (\%) and accuracy gain (\%) on OGC Benchmarks.}
  \label{tbl:mean_prediction_errors_gas} 
\end{table}

\smallskip\noindent\textbf{Measuring The Constraint Violations}
This section simulates the prediction results in an operational 
environment, by measuring the minimum required adjustments in order 
to satisfy the operational limits and the physical constraints in the
energy domains studied. Given the predictions $\hat{y}$ returned by 
a model, the experiments compute a projection $\bar{y}$ of $\hat{y}$ 
into the feasible region and reports the minimal distance $\| \bar{y} 
- \hat{y}\|_2$ of the predictions from the satisfiable solution. 
This step is executed on all the predicted control variables: 
generator dispatch and voltage set points, for power systems, and 
compression ratios, for gas systems. %\footnote{Non-convex least square problems  can be challenging for interior point solvers. 135-pipe is skipped due to convergence issues.} 
Table \ref{tbl:load_flow} reports the minimum distance (normalized 
in percentage) required to satisfy the operational limits and 
physical constraints, and the best results are highlighted in bold. 
These results provide a proxy to evaluate the degree of constraint violations of a model. Notice that the adjustment required decrease with the increase in model complexity. 
\emph{The results show that the LDF can drastically reduce the effort required by a post-processing step to satisfy the 
problem constraints.}

\begin{table}[!tbh]
\centering
\begin{adjustbox}{width=0.8\columnwidth,center}
  \begin{tabular}{l@{\hspace{6pt}} r@{\hspace{2pt}} | @{\hspace{4pt}}
                  r@{\hspace{10pt}} r@{\hspace{10pt}} 
                  l@{\hspace{10pt}} r@{\hspace{10pt}} 
                  l}
  \toprule
  \multirow{2}{*}{\textbf{Test Case}} & 
  \multirow{2}{*}{\textbf{Type}} & 
  ${\cal M}$ &  
  \multicolumn{2}{c}{${\cal M}_\C$} & 
  \multicolumn{2}{c}{${\cal M}_\CD$} \\
  \cmidrule(r){3-3}
  \cmidrule(r){4-5}
  \cmidrule(r){6-7}
  & & violation (\%) & violation (\%) & gain & violation (\%) & gain \\
  \midrule
    \multirow{2}{*}{\textbf{30\_ieee}}
                        & ${p}^g$ & 2.0793 & 0.1815 & (11.45)& \textbf{0.0007} & (2970.0) \\
                        & ${v}$   & 83.138 & 0.0944 & (880.7)& \textbf{0.0037} & (22469) \\
    \multirow{2}{*}{\textbf{118\_ieee}}
                        & ${p}^g$ & 0.1071 & 0.0043 & (24.91)& \textbf{0.0038} & (28.184) \\
                        & ${v}$   & 3.4391 & 0.0956 & (35.97)& \textbf{0.0866} & (39.712) \\
    \multirow{2}{*}{\textbf{300\_ieee}}
                        & ${p}^g$ & 0.0447 & 0.0091 & (4.912)& \textbf{0.0084} & (5.3214) \\
                        & ${v}$   & 31.698 & 0.2383 & (133.0)& \textbf{0.1994} & (158.97) \\
    \midrule
    \multirow{1}{*}{\textbf{24-pipe}}
                        & ${R}$ & 0.1012 & 0.1033 & (0.978)& \textbf{0.0897} & (1.1282) \\
    \multirow{1}{*}{\textbf{40-pipe}}
                        & ${R}$ & 0.0303 & 0.0277 & (1.094)& \textbf{0.0207} & (1.4638) \\
    \multirow{1}{*}{\textbf{135-pipe}}
                        & ${R}$ & 0.0322 & 0.0264 & (1.219)& \textbf{0.0005} & (64.4) \\

    \bottomrule
  \end{tabular}
  \end{adjustbox}
  \caption{Average distances (in percentage) for the active power $p^g$, 
  voltage magnitude $v$, and compressor ratios $R$ of the simulated solutions w.r.t. the corresponding predictions.}
  \label{tbl:load_flow} 
\end{table}

%%%%%%%%%%%%%%%%%%%%%%%%%%%%%%%%%%%%
\subsection{Constrained Predictor Problems}
%%%%%%%%%%%%%%%%%%%%%%%%%%%%%%%%%%%%

This section examines the LDF for constrained predictor problems  discussed in Section \ref{sec:constrained_learning} on transprecision computing and fairness application domains. 

%%%%%%%%%%%%%%%%%%%%%%%%
\smallskip\noindent\textbf{Transprecision computing}
%%%%%%%%%%%%%%%%%%%%%%%%
The benchmark considers training a neural network to predict the error of transprecision configurations. 
The monotonicity property is expressed as a constraint exploiting the relation of dominance among configurations of the train set, i.e. $\nu_i = \max(0, \cM(x_1) - \cM(x_2))$ if $x_1 \preceq x_2$ for every pair $(x_1, x_2)$ in the dataset. 
This approach is particularly suited for instances of training with scarce data points with a high rate of violated constraints, since it guides the learning process towards a more general approximation of the target function.
In order to explore different scenarios the experiments use 5 different 
train sets of increasing size, i.e. 200, 400, 600, 800, and 1000. 
The test set size is fixed to 1000 samples. 
The data sets are constructed by generating random configuration ($d_i$) and 
computing errors ($y$) by measuring the performance loss obtained when 
running the configuration $d_i$ on the target algorithm. 
Ten disjoint training sets are constructed so that the violation constraint 
ratio is $0.5$, while the test set was fixed. 

Table \ref{tbl:SBR} illustrates the average results comparing a model ($\cM$)
that minimizes the Mean Absolute Error (MAE) prediction error, 
one ($\cM_C$) that include the Lagrangian loss functions $\cL_\lambda$ 
associated to each constraint and where all weights $\lambda$ are 
fixed to value $1.0$, and the proposed model ($\cM_C^D$) that uses the LDF to find the optimal Lagrangian weights. 
All prediction model are implemented as classical feed-forward neural 
network with $3$ hidden layers and $10$ units and minimize the MAE as loss function. The training uses 150 epochs, 
Lagrangian step sizes $t_k=10^{-3}$ and learning rate $10^{-3}$. 
The table also show the average number of constraint violations (VC) 
and the sum of the magnitudes of the violated constraints (SMVC), i.e., 
$\sum_{x_i,x_j \in \mathcal{D}; x_i \preceq x_j \land \cM(x_i) > 
\cM(x_j)} |\cM(x_i) - \cM(x_j)|$.

The table clearly illustrates the positive effect of adding the 
constraints within the LDF on reducing the number of constraint violations. 
Notice that model $\cM_C$, that weights all the constraints 
violations equally, produces a degradation of both the MAE score and 
the number and magnitude of the constraint violations, when compared to the baseline model ($\cM$). The benefit of using the LDF is 
substantial in both reducing the number of constraint 
violations and in retaining a high model precision (i.e., a low MAE score).
%finds a good tradeoff between precision and number of 
The most significant contribution was obtained on training sets with 
fewer data points, confirming that \emph{exploiting the Lagrangian 
Duals of the Constraint Violations can be an important tool 
for constrained learning}.

\renewcommand{\B}[1]{\textbf{#1}}

\begin{table}[t]
\begin{adjustbox}{width=0.8\columnwidth,center}
% \begin{tabular}{c|rrrrrrrrr}
\begin{tabular}{l@{\hspace{6pt}} |@{\hspace{6pt}}
               r@{\hspace{10pt}} r@{\hspace{10pt}} r@{\hspace{10pt}} 
               r@{\hspace{10pt}} r@{\hspace{10pt}} r@{\hspace{10pt}}
               r@{\hspace{10pt}} r@{\hspace{10pt}} r@{\hspace{10pt}}} 

\toprule
  $n_{tr}$ & \multicolumn{3}{c}{$\mathcal{M}$} 
  & \multicolumn{3}{c}{$\mathcal{M}_{C}$} & \multicolumn{3}{c}{ $\mathcal{M}^{s}_{D}$} \\
      \cmidrule(r){2-4}
      \cmidrule(r){5-7}
      \cmidrule(r){8-10}
            &     MAE &   VC &    SMVC &       MAE &    VC &    SMVC &     MAE &   VC &    SMVC \\
\midrule
        200 &  0.1902     &  9.6 &  0.2229 &    0.1919 &  35.8 &  0.4748 &  \B{0.1883} &  \B{7.4} &  \B{0.1872} \\
        400 &  0.1765     &  4.5 &  0.0804 &    0.1999 &  19.4 &  0.2149 &  \B{0.1763} &  \B{2.6} &  \B{0.0369} \\
        600 &  \B{0.1687} &  2.5 &  0.0397 &    0.2022 &   9.1 &  0.0683 &  0.1723     &  \B{1.7} &  \B{0.0224} \\
        800 &  \B{0.1672} &  3.0 &  0.0600 &    0.2007 &   8.5 &  0.0746 &  0.1704     &  \B{0.6} &  \B{0.0131} \\
       1000 &  \B{0.1640} &  0.4 &  0.0048 &    0.2012 &   5.7 &  0.0511 &  0.1642     &  \B{0.5} &  \B{0.0043} \\
\bottomrule
\end{tabular}
\end{adjustbox}
  \caption{Mean Absolute Error (MAE), number of constraints 
  violations (VC), and sum of absolute magnitude of violated constraints (SMVC). 
  Best results are highlighted in bold.}
  \label{tbl:SBR} 
\end{table}

%%%%%%%%%%%%%%%%%%%%%%%%
\smallskip\noindent\textbf{Fairness Constraints}
%%%%%%%%%%%%%%%%%%%%%%%%
The benchmark considers building a classifier that minimizes \emph{disparate treatment} \cite{zafar2019}. 
The paper considers the disparate DT index, introduced by 
Aghaei et al.~\cite{AghaeiAV19}, to quantify the disparate impact in a dataset. Given a 
dataset of samples $D = (x_i, y_i)_{i \in [n]}$, this index is defined as:
\begin{align*}
  \textstyle
  \text{DT}(D) = \left| 
  \sum_{x_i \in D_{s_0}} I(\hat{y}_i =1) \right| / \left|D_{s_0}\right| 
  - 
  \left|\sum_{x_i \in D_{s_1}} I(\hat{y}_i =1) \right| / \left|D_{s_1}\right|.
\end{align*}
where $I$ is the characteristic function and $\hat{y}_i$ is the predicted outcome for 
sample $x_i$. The idea is to use a locally weighted average to estimate the conditional expectation. 
The higher is the DT score for a dataset, the more it suffer from disparate treatment, 
with DT $= 0$ meaning that the dataset does not suffer from disparate treatment. 

Since the $DT$ constraint introduced in Equation \eqref{farines_2} is not differentiable with the respect to the model parameters, the paper uses an expectation matching constraints between the predictors for the protected classes, defined as:
\begin{align}
\textstyle
\Big| E_{x \sim D_{s_0}} [ {\cal M}_0(x) | z(x) = 0] -  
      E_{x \sim D_{s_1}} [ {\cal M}_1(x) | z(x) = 1] 
\Big| = 0 
\label{di-score}
\end{align}

The effect of the Lagrangian Dual framework on reducing disparate
treatment was evaluated on three datasets: The \emph{Adult} dataset
\cite{kohavi1996scaling}, containing 30,000 samples and 23 features,
in which the prediction task is that of assessing whether an
individual earns more than $50K$ per year and the protected attribute is \emph{race}. The \emph{Default} of Taiwanese credit card users
\cite{yeh2009comparisons}, containing 45,000 samples and 13 features,
in which the task is to predict whether an individual will default and the protected attribute is \emph{gender}. Finally, the \emph{Bank} dataset \cite{zafar2019}, containing 41,188 samples, each with 20 attributes, where the task is to predict whether an individual has subscribed or not and the protected attribute is \emph{age}.  The experiments use a 80/20 train/test split and executes a 5-fold  cross-validation to evaluate the accuracy and the fairness score (DT) of the predictors. 

Table \ref{tbl:fairness} illustrates the results comparing model
$\cM$ that minimizes the Binary Cross Entropy (BCE) loss, model
$\cM_C$ that includes the Lagrangian loss functions $\cL_\lambda$
associated with each constraint and where all $\lambda$ are fixed to
value $1.0$, and the proposed model $\cM_C^D$ that uses the LDF to find the optimal Lagrangian weights.  All prediction models use a classical feed-forward neural network with  $3$ layers and $10$ hidden units. The training uses 100 epochs, Lagrangian step size $s_k=10^{-4}$ and learning rate $10^{-3}$. 
The models are also compared against a state-of-the-art fair classifier which enforces fairness by limiting the covariance between the loss function and the sensitive variable \cite{zafar2019}. While \cite{zafar2019} focuses on logistic regression, the model is implemented as a neural network with the same hyper parameters of model $\cM$. 
% However, when the true decision boundary for the classification task is highly non-linear a simple \textbf{LR} might not  compete well with non-linear classifiers such as deep networks which our models are based on. 
The table clearly shows the effect of the Lagrangian
constraints on reducing the DT score. Not only such reduction attains state-of-the-art results on the DT score, but it also comes at a much more contained cost of accuracy degradation. 

\begin{table}[!t]
\centering
\begin{adjustbox}{width=0.8\columnwidth,center}
  \begin{tabular}{l@{\hspace{6pt}} |@{\hspace{6pt}}
                 r@{\hspace{10pt}} r@{\hspace{10pt}} 
                 r@{\hspace{10pt}} r@{\hspace{10pt}} 
                 r@{\hspace{10pt}} r@{\hspace{10pt}}
                 r@{\hspace{10pt}} r@{\hspace{10pt}}} 
     \toprule
     \multirow{2}{*}{\textbf{Dataset}}&  
     \multicolumn{2}{c}{$\mathcal{M}$}&
     \multicolumn{2}{c}{$\mathcal{M}_C$}&
     \multicolumn{2}{c}{$\mathcal{M}_C^D$}&
     \multicolumn{2}{c}{Zafar'19}\\
      \cmidrule(r){2-3}
      \cmidrule(r){4-5}
      \cmidrule(r){6-7}
      \cmidrule(r){8-9}
     & Acc. &  DT & Acc. &  DT & Acc. &  DT & Acc. &  DT\\
     \midrule
     Adult  & \textbf{0.8423} & 0.1853 
            & 0.8333 & 0.0627
            & 0.8335 & \textbf{0.0545} 
            & 0.7328 & 0.1037\\
     Default  &0.8160 & 0.0162
              &\textbf{0.8182} & 0.0216
              &0.8166 & \textbf{0.0101}
              &0.6304 & 0.0109\\
     Bank   &\textbf{0.8257} & 0.4465
            &0.7744 & 0.4515
            &0.8135& 0.1216
            &0.7860 & \textbf{0.0363} \\
   \bottomrule
  \end{tabular}
  \end{adjustbox}
  \caption{Classification accuracy (Acc.) and fairness score (DT)}
  \label{tbl:fairness} 
\end{table}

\section{Related Work}
The application of Deep Learning to constrained optimization problems is receiving increasing attention. 
Approaches which embed optimization components in neural networks include \cite{pointer_nets,khalil2017learning,kool2018attention}. 
These approaches typically rely on problems exhibiting properties like convexity or submodularity. 
Another line of work leverages explicit optimization algorithms as a differentiable layer into neural networks \cite{amos2017optnet,donti2017task,wilder2019melding}. 
A further collection of works interpret constrained optimization as a two-player game, in which one player optimizes the objective function and a second player attempt at satisfying the problem constraints \cite{kearns2017,narasimhan2018,agarwal2018}. 
For instance Agarwal et al.~\cite{agarwal2018}, proposes a best-response algorithm applied to fair classification for a class of linear fairness constraints.
To study generalization performance of training algorithms that learn to satisfy the problem constraints, Cotter et al.~\cite{cotter2018} propose a two-players game approach in which one player optimizes the model parameters on the training data and the other player optimizes the constraints on a validation set. 
Arora et al.~\cite{arora2012multiplicative} proposed the use of a multiplicative rule to iteratively changing the weights of different distributions to maintaining some properties and discuss the applicability of the approach to a constraint satisfaction domain. 

A different strategy for minimizing empirical risk subject to a set of constraints is that of using projected stochastic gradient descent (PSGD). Cotter et al.~\cite{cotter2016} proposed an extension of PSGD that stay close to the feasible region while applying constraint probabilistically at each iteration of the learning cycle. 

Different from these proposal, this paper proposes a framework that exploits key ideas in Lagrangian duality to encourage the satisfaction of generic constraints within a neural network learning cycle and apply to both sample dependent constraints (as in the case of energy problems) and dataset dependent constraints (as in the case of transprecision computing and fairness problems). This paper builds on the recent results that were dedicated to learning and optimization in power systems \cite{ferdin2020predicting}. %It proposes a Lagrangian Dual Framework that 

\section{Conclusions}

This paper proposed a Lagrangian dual framework to encourage the satisfaction of constraints in deep learning. It was motivated by a desire to learn parametric constrained optimization problems that feature complex physical and engineering constraints. The paper showed how to exploit Lagrangian duality for deep learning to obtain predictors that minimize constraint violations. The proposed framework can be applied to constrained optimization problems, in which the constraints model relations among features of each data sample, and to constrain predictors in which the constraints enforce global properties over multiple dataset samples and the predictor outputs. 
%Moreover, it showed that the proposed approach can be applied to constrained learning problems where the learning task imposes constraints on the predictor itself. 

The Lagrangian dual framework for deep learning was evaluated on a collection of realistic energy networks, by enforcing non-discriminatory decisions on a variety of datasets, and on a transprecision computing application. The results demonstrated the effectiveness of the proposed method that dramatically decreases constraint violations %(e.g. up to 80\% in transprecision computing) 
committed by the predictors and, in some applications, as in those in energy optimization, increases the prediction accuracy by up to two orders of magnitude.

% \noindent\textbf{Reproducibility}: Datasets and model used will be released upon publication.

\bibliographystyle{plain}
\bibliography{dl_opf}

\appendix

\section{Energy System Case Studies}

This section extends the description of the energy applications for evaluating the Lagrangian Dual Framework: Optimal power flow in electricity networks and the Optimal compressor controls in gas networks. Both problems are nonlinear and nonconvex. 

\begin{model}[t]
  {\small
  \caption{${\cal O}_{\text{OPF}}$: AC Optimal Power Flow}
  \label{model:ac_opf}
  \vspace{-6pt}
  \begin{align}
    \mbox{\bf variables:} \;\;
    & S^g_i, V_i \;\; \forall i\in N, \;\;
      S^f_{ij}   \;\; \forall(i,j)\in E \cup E^R \nonumber \\
    \mbox{\bf minimize:} \;\;
    & {\cO}(\bm{S^d}) = \sum_{i \in N} {c}_{2i} (\Re(S^g_i))^2 + {c}_{1i}\Re(S^g_i) + {c}_{0i} \label{ac_obj} \\
    \mbox{\bf subject to:} \;\; 
    & \angle V_{i} = 0, \;\; i \in N \label{eq:ac_0} \\
    & {v}^l_i \leq |V_i| \leq {v}^u_i     \;\; \forall i \in N \label{eq:ac_1} \\
    & {\theta}^{l}_{ij} \leq \angle (V_i V^*_j) \leq {\theta}^{u}_{ij} \;\; \forall (i,j) \in E  \label{eq:ac_2}  \\
    & {S}^{gl}_i \leq S^g_i \leq {S}^{gu}_i \;\; \forall i \in N \label{eq:ac_3}  \\
    & |S^f_{ij}| \leq {s}^{fu}_{ij}          \;\; \forall (i,j) \in E \cup E^R \label{eq:ac_4}  \\
    & S^g_i - {S}^d_i = \textstyle\sum_{(i,j)\in E \cup E^R} S^f_{ij} \;\; \forall i\in N \label{eq:ac_5}  \\ 
    & S^f_{ij} = {Y}^*_{ij} |V_i|^2 - {Y}^*_{ij} V_i V^*_j       \;\; \forall (i,j)\in E \cup E^R \label{eq:ac_6}
  \end{align}
  }
  \vspace{-12pt}
\end{model}

\subsection{Optimal Power Flow} \emph{Optimal Power Flow (OPF)} is
the problem of finding the best generator dispatch to meet the demands
in a power network, while satisfying challenging transmission constraints 
such as the nonlinear nonconvex AC power flow equations and also operational limits such as voltage and generation bounds.
Finding good OPF predictions are important, as a 5\% reduction in generation costs could save billions of dollars (USD) per year~\cite{Cain12historyof}.
In addition, the OPF problem is a fundamental building bock of many
applications, including security-constrained OPFs \cite{monticelli:87}), optimal transmission switching \cite{OTS}, capacitor placement \cite{baran:89}, and expansion planning \cite{verma:16}.

Typically, generation schedules are updated in intervals of 5
minutes \cite{Tong:11}, possibly using a solution to the OPF solved in
the previous step as a starting point. In recent years, the
integration of renewable energy in sub-transmission and distribution
systems has introduced significant stochasticity in front and behind
the meter, making load profiles much harder to predict and introducing
significant variations in load and generation. This uncertainty forces
system operators to adjust the generators setpoints with increasing
frequency in order to serve the power demand while ensuring stable
network operations. However, the resolution frequency to solve OPFs is
limited by their computational complexity. To address this issue,
system operators typically solve OPF approximations such as the linear
DC model (DC-OPF).  While these approximations are more efficient
computationally, their solution may be sub-optimal and induce
substantial economical losses, or they may fail to satisfy the
physical and engineering constraints.

Similar issues also arise in expansion planning and other
configuration problems, where plans are evaluated by solving a massive
number of multi-year Monte-Carlo simulations at 15-minute intervals
\cite{pachenew,Highway50}. Additionally, the stochasticity introduced
by renewable energy sources further increases the number of scenarios
to consider.  Therefore, modern approaches recur to the linear DC-OPF
approximation and focus only on the scenarios considered most
pertinent \cite{pachenew} at the expense of the fidelity of the
simulations.

A power network $\bm{\cN}$ can be represented as
a graph $(N, E)$, where the nodes in $N$ represent buses and the edges
in $E$ represent lines. The edges in $E$ are directed and $E^R$ is
used to denote those arcs in $E$ but in reverse direction.  The AC
power flow equations are based on complex quantities for current $I$,
voltage $V$, admittance $Y$, and power $S$, and these equations are a
core building block in many power system applications.
Model~\ref{model:ac_opf} shows the AC OPF formulation, with
variables/quantities shown in the complex domain.  Superscripts $u$ and
$l$ are used to indicate upper and lower bounds for variables. The
objective function ${\cO}(\bm{S^g})$ captures the cost of the
generator dispatch, with $\bm{S^g}$ denoting the vector of generator
dispatch values $(S^g_i \:|\: i \in N)$.  Constraint \eqref{eq:ac_0}
sets the reference angle to zero for the slack bus $i \in N$ to
eliminate numerical symmetries.  Constraints \eqref{eq:ac_1} and
\eqref{eq:ac_2} capture the voltage and phase angle difference bounds.
Constraints \eqref{eq:ac_3} and \eqref{eq:ac_4} enforce the generator
output and line flow limits.  Finally, Constraints \eqref{eq:ac_5}
capture Kirchhoff's Current Law and Constraints \eqref{eq:ac_6}
capture Ohm's Law.

Table~\ref{tbl:power_dataset} describes the power network benchmarks used in main text,
including the number of buses $|{\cal N}|$, transmission lines/transformers $|{\cal E}|$,
loads $l$, and generators $g$. 
\begin{table}[t]
\centering
\resizebox{0.35\textwidth}{!}
{
  \begin{tabular}{l@{\hspace{6pt}} |c@{\hspace{6pt}} 
                 c@{\hspace{6pt}} c@{\hspace{6pt}} c@{\hspace{6pt}}}
  \toprule
    \multicolumn{1}{l}{\textbf{Test Case}} &  $|{\cal N}|$ & $|{\cal E}|$ & $l$ & $g$ \\
    \midrule
  \textbf{30\_ieee     }& 30& 82&  21 & 2    \\
  \textbf{118\_ieee    }& 118& 372& 99 & 19  \\
  \textbf{300\_ieee    }& 300& 822& 201& 57  \\
  \bottomrule
  \end{tabular}
}
\caption{The Power Networks DataSet.}
% Number of buses, 
%Number of directed transmission lines}
\label{tbl:power_dataset} 
\end{table}

\subsection{Optimal Compressor Optimization}

\emph{Optimal Gas Flow (OGF)} is the problem of finding the best
compression control to maintain pressure requirements in a natural gas
pipeline system. 
Similar to the OPF problem, the gas flow 
problem is a non-convex non-linear optimization problem, with 
challenging nonconvex function to measure the costs of compressors.
It is also a fundamental building block for many gas pipeline problems,
including: 
gas pipeline expansion planning~\cite{borrazsanchez16convex},
dynamic compressor optimization~\cite{mak19dynamic}, and 
joint gas-grid transmission planning problem~\cite{bent18joint}.

Historically, natural gas demands came from utilities or large industrial
customers whose demands are predictable with little variations. 
These demands are often traded using day-ahead contracts.
Therefore, operators can often assume that injections and withdrawals would be similar to the past
and re-use precomputed control set-points. 
In the past decade, the increasing penetration of renewable energies into power systems 
has driven an increase in installations of gas-powered electric generators. 
Gas-powered generators can start up and shut down several times a day, and also 
capable to rapidly adjust their production to balance 
the fluctuation of renewable energy sources.
However, the growing use of gas-powered generators 
implies substantial intra-day high-volume gas fluctuations, and 
has prompted concerns within the industries.
Similar to OPF, the resolution frequency to solve OGF problem
is limited by the computational complexity of the system,
and system operators typically solve OGF approximations instead 
such as reduced-order models~\cite{zlotnik15optimal} or convex relaxations~\cite{borrazsanchez16convex}.
These approximations are more efficient computationally, but 
can be sub-optimal and/or fail to capture the physical and operational limits.

\begin{model}[t]
  {\small
  \caption{${\cal O}_{\text{OGF}}$: Optimal Gas Flow}
  \label{model:ac_ogf}
  \vspace{-6pt}
  \begin{align}
    \mbox{\bf variables:} \;\;
    & p_i, q_i \; \forall i \in \cJ, \;
      q_{ij} \; \forall (i,j) \in \cP, \;
      R_{ij} \; \forall(i,j) \in \cC \nonumber \\
    \mbox{\bf minimize:} \;\;
    & {\cO}(\bm{q}) = \sum_{(i,j) \in C} \mu^{-1}|q_{ij}|(\max\{R_{ij},1\}^{2(\gamma-1)/\gamma}-1) \label{gas_obj} \\
    \mbox{\bf subject to:} \;\; 
    & \textstyle\sum_{(i,j) \in \cP} q_{ij} - \sum_{(j,i) \in \cP} q_{ji} = q_i, \;\; \forall i \in \cJ \label{eq:gas_0} \\
    & {p_i}^l \leq p_i \leq {p_i}^u     \;\; \forall i \in N, \;\; {q_{ij}}^l \leq q_{ij} \leq {q_{ij}}^u     \;\; \forall (i,j) \in \cP \label{eq:gas_1} \\
    & {R_{ij}}^l \leq R_{ij} \leq {R_{ij}}^u    \;\; \forall (i,j) \in \cC \label{eq:gas_2}\\
    & p_i = p^T_i   \;\; \forall i \in \cJ^B, q_i = 0   \;\; \forall i \in \cJ^T, q_i = q^d_i \;\; i  \in \cJ^D\label{eq:gas_4}\\
    & R_{ij}^2 p_{i}^2 - p_{j}^2 = L_{ij} \displaystyle \frac{\lambda a^2}{D_{ij} A_{ij}^2} q_{ij} \lvert q_{ij} \rvert    \;\; \forall (i,j)\in \cC \label{eq:gas_5}\\
    & p_{i}^2 - p_{j}^2 = L_{ij} \displaystyle \frac{\lambda a^2}{D_{ij} A_{ij}^2} q_{ij} \lvert q_{ij} \rvert     \;\; \forall (i,j)\in \cP - \cC \label{eq:gas_6}
  \end{align}
  }
  \vspace{-8pt}
\end{model}

A natural gas network can be represented as a
directed graph $\bm{\cN}=(\cJ,\cP)$, where a node $i \in \cJ$
represents a junction point and an arc represents a pipeline $(i,j)
\in \cP$ represent the edges.  Compressors ($\cC \subseteq \cP$) are
installed in a subset of the pipelines for boosting the gas pressure
$p$ in order to maintain pressure requirements for gas flow $q$.  The
set $\cJ^D$ of gas demands and the set $\cJ^T$ of transporting nodes
are modeled as junction points, with net gas flow $q_i$ set to the gas
demand ($q^d_i$) and zero respectively.  For simplicity, the paper
assumes no pressure regulation and losses within junction nodes and
gas flow/flux are conserved throughout the system.  A subset $\cJ^B
\in \cJ$ of the nodes are regulated with constant pressure $p^T_i$.
The length of pipe $(i,j)$ is denoted by $L_{ij}$, its diameter by
$D_{ij}$, and its cross-sectional area by $A_{ij}$.  Universal
quantities include isentropic coefficient $\gamma$, compressor
efficiency factor $\mu$, sound speed $a$, and gas friction factor
$\lambda$.  Model~\ref{model:ac_ogf} depicts the OGF formulation.  The
objective function ${\cO}(\bm{q})$ captures the compressor costs using
the compressor control values $(R_{ij} \:|\: (i,j) \in \cC)$.
Constraints \eqref{eq:gas_0} capture the flow conversation equations.
Constraints \eqref{eq:gas_1} and \eqref{eq:gas_2} capture the
pressure, flux flow, and compressor control bounds.  Constraints
\eqref{eq:gas_4} set the boundary conditions for the demands and the
regulated pressures. Finally, Constraints \eqref{eq:gas_5} and
\eqref{eq:gas_6} capture the steady-state isothermal gas flow
equation.

Table~\ref{tbl:gas_dataset} describes the gas network benchmarks used in the main text,
including the number of junctions $|{\cJ}|$, pipelines $|{\cP}|$,
compressors $|\cC|$, and active gas loads $|\cJ^D|$. 
\begin{table}[t]
\centering
\resizebox{0.35\textwidth}{!}
{
  \begin{tabular}{l@{\hspace{6pt}} |c@{\hspace{6pt}} 
                 c@{\hspace{6pt}} c@{\hspace{6pt}} c@{\hspace{6pt}}}
  \toprule
    \multicolumn{1}{l}{\textbf{Test Case}} &  $|{\cJ}|$ & $|{\cP}|$ & $|{\cC}|$ & $|\cJ^D|$ \\
    \midrule
  \textbf{24-pipe      }& 25  & 24   & 5 &  8 \\
  \textbf{40-pipe      }& 40  & 45   & 6 &  26 \\
  \textbf{135-pipe     }& 135 & 170  & 10&  19 \\
  \bottomrule
  \end{tabular}
}
\caption{The Gas Networks DataSet.}
% Number of buses, 
%Number of directed transmission lines}
\label{tbl:gas_dataset} 
\end{table}

\section{Learning Model Details}
% Also, we need the network information -- also from the AAAI extended version.
% }
This section describes the modeling details of the proposed LDF related to the energy applications described in the previous section. 

\subsection{Lagrangian relaxation}
This paper uses a Lagrangian
relaxation approach based on constraint violations
\cite{Fontaine:14} used in generalized augmented Lagrangian
relaxation \cite{Hestenes:69}. 
To fully describe the modeling details 
for the OPF and OGF problems, an optimization problem {\cal O} definition will be slightly expanded as:
\begin{flalign*}
\mbox{\bf minimize:} & \;\; f(\by) \\
\mbox{\bf subject to:} & \;\; h(\by) = 0\\ 
                       &\;\; g(\by) \leq 0 
\end{flalign*}
where equality constraints $h$ and inequality constraints $g$
are separated.
% \begin{flalign*}
% \mbox{\bf minimize:} \; f(\bx) \;\;
% \mbox{\bf subject to:} \; h(\bx) = 0; \;\;
%                           g(\bx) \leq 0 
% \end{flalign*}
The Lagrangian relaxation of {\cal O} is given by
\begin{flalign*}
\mbox{\bf minimize:} & \;\; f(\by) + \lambda_h h(\by) + \lambda_g g(\by)
\end{flalign*}
\noindent
where $\lambda_h$ and $\lambda_g \geq 0$ are the Lagrangian multipliers
for the equality constraints and inequality constraints.
In contrast, the violation-based Lagrangian relaxation is 
\begin{flalign*}
\mbox{\bf minimize:} & \;\; f(\by) + \lambda_h |h(\by)| + \lambda_g \max(0,g(\by))
\end{flalign*}
\noindent
with $\lambda_h,\lambda_g \geq 0$. In other words, the traditional
Lagrangian relaxation exploits the satisfiability degrees of
constraints, while the violation-based Lagrangian relaxation is
expressed in terms of violation degrees. 
The satisfiability degree of an inequality constraint measures how well the
constraint is satisfied, with negative values representing the slack
and positive values representing violations, while the violation 
degree is always non-negative and represents how much the  constraint is 
violated.
More formally, the satisfiability degree of a constraint 
$c \!:\! \mathbb{R}^n \to \text{Bool}$
is a function $\sigma_c\!:\! \mathbb{R}^n \to \mathbb{R}$ such that
$\sigma_c(\by) \leq 0 \equiv c(\by)$. The violation degree of a
constraint $c\!:\! \mathbb{R}^n \to \text{Bool}$ is a function
$\nu_c\!:\! \mathbb{R}^n \to \mathbb{R^+}$ such that 
$\sigma_c(\by) \equiv 0 \equiv c(\by)$. 
For instance, for a linear constraints $c(\by)$ of
type $A\by \geq b$, the \emph{satisfiability degree} is defined as
\begin{equation*}
\sigma_c(\by) \equiv \bm{b} - A\by
\end{equation*}
and the \emph{violation degrees} for inequality and equality 
constraints are specified by
\begin{equation*}
\nu^{\geq}_c(\by) = \max\left(0, \sigma_c(\by)\right) 
\qquad
\nu^{=}_c(\by) = \left| \sigma_c(\by) \right|. 
\end{equation*}
\noindent
Although the resulting term is not differentiable (but admits
subgradients), computational experiments indicated that violation
degrees are more appropriate for prediction than satisfiability
degrees. 
%Observe also that an augmented Lagrangian method uses both
%the satisfiability and violation degrees in its objective.
Even though in theory all the constraints can be naively incorporated into 
the learning model by constructing the violation degree metric functions, 
in practice not all the constraints are needed.
For example, if a constraint is guaranteed to be satisfied by the input data (e.g. (13) in Model~\ref{model:ac_ogf}),
the constraint can be omitted as the violation degree is always zero.
On the other hand, if the satisfiability of a constraint depends on
the prediction or the constraint is used to compute an indirect prediction (e.g. Ohm's Law (8) in Model~\ref{model:ac_opf}), the violation degree of the constraint can be measured directly against the ground truth. 

\subsection{OPF satisfiability and violation degrees}
\label{sec:opf_details}
% To define the violation degrees of the AC-OPF constraints, the
% baseline model needs to extended to predict the reactive power
% dispatched $\bm{q}^g$ and the voltage angles $\bm{\theta}$ of the
% power network.  
Given the predicted values:
$\hat{\bm{V}} = \hat{\bm{v}} \angle \hat{\bm{\theta}} $ for voltages,
$\hat{\bm{S}}^g = \hat{\bm{p}}^g + i\hat{\bm{q}}^g$ for generation dispatches,
and 
$\hat{\bm{S}}^f = \hat{\bm{p}}^f + i\hat{\bm{q}}^f$ for lines/transformers flows,
this section extends the main paper by 
reporting the complete set of satisfiability $\sigma(\cdot)$ and violation  degrees $\nu(\cdot)$ for the OPF problem. 
% Given the predicted values $\hat{\bm{v}}, \hat{\bm{\theta}}, 
% \hat{\bm{p}}^g,$ and $\hat{\bm{q}}^g$, the satisfiability degree of 
% the OPF constraints are expressed as follows:
\begin{flalign*}
\sigma_{3a}(\hat{v}_i) &= 
    {v}^{l}_i - \hat{v}_i &\!\!\!\!\!\!
    \forall i \in {N} && 
\sigma_{3b}(\hat{v}_i) &=  
    \hat{v}_i - {v}^{u}_i &\!\!\!\!\!\! \forall i \in {N}\\
\sigma_{4a}(\hat{\theta}_{ij}) &=  
    (\hat{\theta}_j - \hat{\theta}_i) - {\theta}^{\Delta}_{ij} &\!\!\!\!\!\! \forall (ij) \in { E} &&
\sigma_{4b}(\hat{\theta}_{ij}) &= 
    (\hat{\theta}_i - \hat{\theta}_j) -{\theta}^{\Delta}_{ij} &\!\!\!\!\!\! \forall (ij) \in { E}\\
\sigma_{5a}(\hat{p_i}^g) &= 
    {p}^{gl}_i - \hat{p}{}^g_i &\!\!\!\!\!\! \forall i \in {N}&&
\sigma_{5b}(\hat{p_i}^g) &=  
    \hat{p}^g_i - {p}^{gu}_i &\!\!\!\!\!\! \forall i \in {N}\\
\sigma_{5c}(\hat{q_i}^g) &= 
    {q}^{gu}_i - \hat{q}{}^g_i &\!\!\!\!\!\! \forall i \in {N}&&
\sigma_{5d}(\hat{q_i}^g) &=  
    \hat{q}^g_i - {q}^{gu}_i &\!\!\!\!\!\! \forall i \in {N}\\
\sigma_6(\hat{p}^f_{ij}, \hat{q}^f_{ij}) &= 
    (\hat{p}^f_{ij})^2 + (\hat{q}^f_{ij})^2 - {s}^{u}_{ij} &\!\!\!\!\!\! \forall (ij) \in {E} \cup E^R\\
%%%%%%%%%%
\sigma_{7a}(\hat{p}^g_i, {p}^d_i, \hat{\bm{p}}^f) &= 
      \sum_{(ij)\in {E}} \hat{p}^f_{ij} - (\hat{p}^g_i \!-\! {p}^d_i)
      &\!\!\!\!\!\! \forall i \in {N}\\
\sigma_{7b}(\hat{q}^g_i, {q}^d_i, \hat{\bm{q}}^f) &= 
      \sum_{(ij)\in {E}} \hat{q}^f_{ij} - (\hat{q}^g_i \!-\! {q}^d_i)
      &\!\!\!\!\!\! \forall i \in {N}\\
\sigma_{8a}(\hat{{p}}^f_{ij}, {p}^f_{ij}) &= 
    \hat{p}^f_{ij} - p_{ij}^f &\!\!\!\!\!\! \forall (ij) \in {E}&&
\sigma_{8b}(\hat{{q}}^f_{ij}, {q}^f_{ij}) &= 
    \hat{q}^f_{ij} - q^f_{ij} &\!\!\!\!\!\! \forall (ij) \in {E}
\end{flalign*}
Functions $\sigma_{3a}$ and $\sigma_{3b}$ correspond to Constraints (3) 
and capture the distance of the voltage predictions $\hat{v}_i$ 
and its bounds. 
Similarly, $\sigma_{4a}$ and $\sigma_{4b}$ correspond to 
(4) and measure the difference between 
voltage phase angle differences and its bound. 
Functions $\sigma_{5a}$ to $\sigma_{5d}$ 
correspond to (5) and describe the distance of the 
generator active and reactive dispatch predictions from their upper and lower bounds. 
$\sigma_6$ corresponds to (6) and 
measures distance between the squared apparent power and its bound. 
%Therein, $\tilde{p}^f_{ij}$ and $\tilde{q}^f_{ij}$ are, respectively,
% the active and reactive power flow for line $(ij) \in {\cal E}$. 
Functions $\sigma_{7a}$ and $\sigma_{7b}$ relate to the 
Kirchhoff Current Law (7)  
and measure the power flow violations at every bus (i.e. bus injection violations).
Finally, functions $\sigma_{8a}$ and $\sigma_{8b}$ measure the deviation of 
the indirect predicted flows 
$\hat{p}^f_{ij}$ and $\hat{q}^f_{ij}$
to their ground truth values
${p}^f_{ij}$ and ${q}^f_{ij}$.
Note that $\hat{p}^f_{ij}$ and $\hat{q}^f_{ij}$ are not direct
predictions from the output of the DNN. These quantities are
indirect predictions computed using the voltage predictions.
%%%%%%

% For an inequality constraint indexed $\iota$, the satisfiability degree associates two quantities: $\sigma_\iota L$ and $\sigma_\iota R$, capturing, respectively, the two sides of the inequality. 
The violation degrees associated to the satisfiability degree 
above are defined below.
{\small
\begin{flalign*}
\nu_{3}(\hat{\bm{v}}) = 
    \frac{1}{\lvert N \rvert} \sum_{i \in {N}}
    \left(\nu_c^{\geq}\big(\sigma_{3a}(\hat{v}_i)\big) + 
    \nu_c^{\geq}\big(\sigma_{3b}(\hat{v}_i)\big) \right)\\
\nu_{4}(\hat{\bm{\theta}}) = 
    \frac{1}{\lvert E \rvert} \sum_{(ij) \in {E}}
    \left(\nu_c^{\geq}\big(\sigma_{4a}(\hat{\theta}_{ij})\big) + 
    \nu_c^{\geq}\big(\sigma_{4b}(\hat{\theta}_{ij})\big) \right)\\
\nu_{5a}(\hat{\bm{p}}^g) = 
    \frac{1}{\lvert N \rvert} \sum_{i \in {N}}
    \left(\nu_c^{\geq}\big(\sigma_{5a}(\hat{p}_i^g)\big) + 
    \nu_c^{\geq}\big(\sigma_{5b}(\hat{p}_i^g)\big) \right)&&
\nu_{5b}(\hat{\bm{q}}^g) = 
    \frac{1}{\lvert N \rvert} \sum_{i \in {N}}
    \left(\nu_c^{\geq}\big(\sigma_{5c}(\hat{q}_i^g)\big) + 
    \nu_c^{\geq}\big(\sigma_{5d}(\hat{q}_i^g)\big) \right)\\
\nu_{6}(\hat{\bm{p}}^f\!, \hat{\bm{q}}^f) = 
    \frac{1}{\lvert E \rvert} \sum_{(ij) \in {E}}
    \nu_c^{\geq}\big(\sigma_{6}(\hat{p}_{ij}^f, \hat{q}_{ij}^f)\big)\\
\nu_{7a}(\hat{\bm{p}}^g\!, {\bm{p}}^d\!, \bm{p}^f) = 
    \frac{1}{\lvert E \rvert} \sum_{(ij) \in {E}}
    \nu_c^{=}\big(\sigma_{7a}(\hat{p}^g_i, {p}^d_i, \tilde{\bm{p}}^f)\big)&&
\nu_{7b}(\hat{\bm{q}}^g\!, {\bm{q}}^d\!, \bm{q}^f) = 
    \frac{1}{\lvert E \rvert} \sum_{(ij) \in {E}}
    \nu_c^{=}\big(\sigma_{7b}(\hat{q}^g_i, {q}^d_i, \tilde{\bm{q}}^f)\big)\\
\nu_{8a}(\hat{\bm{p}}^f\!, \bm{p}^f) = 
    \frac{1}{\lvert E \rvert} \sum_{(ij) \in {E}}
    \nu_c^{=}\big(\sigma_{8a}(\hat{p}_{ij}^f, {p}_{ij}^f)\big)&&
\nu_{8b}(\hat{\bm{q}}^f\!, \bm{q}^f) = 
    \frac{1}{\lvert E \rvert} \sum_{(ij) \in {E}}
    \nu_c^{=}\big(\sigma_{8b}(\hat{q}_{ij}^f, {q}_{ij}^f)\big).
\end{flalign*}
}
These functions capture the average deviation by which the
prediction violates the associated constraint. 
The violations degrees define 
penalties that will be used to enrich the DNN loss function to encourage their 
satisfaction.  

\subsection{OGF satisfiability and violation degrees}
Given the predicted values:
$\hat{\bm{R}}$ for compression controls,
$\hat{\bm{p}}$ for pressure values,
$\hat{\bm{q}}$ for natural gas supply,
and 
$\hat{\bm{q}}^f$ for pipelines flows,
this section extends the main paper by 
reporting the complete set of satisfiability $\sigma(\cdot)$ and violation 
degrees $\nu(\cdot)$ for the OGF problem. 
\begin{flalign*}
\sigma_{10}(\hat{\bm{q}}^f, \hat{q}_i) &= 
    \sum_{(i,j) \in \cP} \hat{q}^f_{ij} - \sum_{(j,i) \in \cP} \hat{q}^f_{ji} -\hat{q}_i &\!\!\!\!\!\!
    \forall i \in \cJ\\
\sigma_{11a}(\hat{p}_i) &= 
    p^l_i - \hat{p}_i &\!\!\!\!\!\!
    \forall i \in \cJ &&
\sigma_{11b}(\hat{p}_i) &= 
    \hat{p}_i - p^u_i  &\!\!\!\!\!\!
    \forall i \in \cJ\\
\sigma_{11c}(\hat{q}^f_{ij}) &= 
    q^l_{ij} - \hat{q}_{ij} &\!\!\!\!\!\!
    \forall (i,j) \in \cP &&
\sigma_{11d}(\hat{q}^f_{ij}) &= 
    \hat{q}_{ij} - q^u_{ij}  &\!\!\!\!\!\!
    \forall (i,j) \in \cP\\
\sigma_{12a}(\hat{R}_{ij}) &= 
    R^l_{ij} - \hat{R}_{ij}   &\!\!\!\!\!\!
    \forall (i,j) \in \cC &&
\sigma_{12b}(\hat{R}_{ij}) &= 
    \hat{R}_{ij} - R^u_{ij}  &\!\!\!\!\!\!
    \forall (i,j) \in \cC\\
% \sigma_{14}(\hat{\bm{R}}, \hat{\bm{p}},\hat{\bm{q}}^f) &= 
%      \hat{R}_{ij}^2 \hat{p}_{i}^2 - \hat{p}_{j}^2 - L_{ij} \frac{\lambda a^2}{D_{ij} A_{ij}^2} \hat{q}^f_{ij} \lvert \hat{q}^f_{ij} \rvert &\!\!\!\!\!\!
%     \forall (i,j) \in \cC\\
% \sigma_{15}(\hat{\bm{p}},\hat{\bm{q}}^f) &= 
%      \hat{p}_{i}^2 - \hat{p}_{j}^2 - L_{ij} \frac{\lambda a^2}{D_{ij} A_{ij}^2} \hat{q}^f_{ij} \lvert \hat{q}^f_{ij} \rvert &\!\!\!\!\!\!
%     \forall (i,j) \in \cP - \cC\\
\sigma_{14}(\hat{q}^f_{ij}, q^f_{ij}) &= 
     \hat{q}^f_{ij} - {q}^f_{ij} &\!\!\!\!\!\!
    \forall (i,j) \in \cP\\    
\end{flalign*}
% \terrence{I believe there should be no satisfiability degree functions and violation degree functions for the pressure loss constraints (similar to OPF), i.e. (14) and (15) in the OGF model (see commented parts above). Nando ... please check your implementation.}
Function $\sigma_{10}$ corresponds to Constraints (10) 
and measure the gas flow violations at every junction.
Functions $\sigma_{11a} - \sigma_{11d}$ relate to Constraints (11)
and capture the distance of the pressure and gas flow predictions $\hat{p}_i$
and $\hat{q}_{ij}$ to their bounds. 
Similarly, $\sigma_{12a}$ and $\sigma_{12b}$ correspond to Constraints (12)
and capture the distance of the compression control $\hat{R}_{ij}$ to its bounds.
Finally, functions $\sigma_{14}$ measure the 
deviation of the indirect predicted flows $\hat{q}^f_{ij}$ to its
ground truth value $q^f_{ij}$. 
Similar to OPF, $\hat{q}^f_{ij}$ are not direct predictions from the output
of the DNN and computed using the pressure and compressor control predictions. 
Note that Constraints (13) are skipped since  

The violation degrees associated to the satisfiability degree 
above are defined below.
{\small
\begin{flalign*}
\nu_{10}(\hat{\bm{q}}^f\!, \hat{\bm{q}}) &= 
    \frac{1}{\lvert \cJ \rvert} \sum_{i \in \cJ}
    \nu_c^{=}\big(\sigma_{10}(\hat{\bm{q}}^f, \hat{q}_i)\big)\\
\nu_{11a}(\hat{\bm{p}}) &= 
    \frac{1}{\lvert \cJ \rvert} \sum_{i \in \cJ}
    \left(\nu_c^{\geq}\big(\sigma_{11a}(\hat{p}_i)\big) + 
    \nu_c^{\geq}\big(\sigma_{11b}(\hat{p}_i)\big) \right)\\
\nu_{11b}(\hat{\bm{q}}^f) &= 
    \frac{1}{\lvert \cP \rvert} \sum_{i \in P}
    \left(\nu_c^{\geq}\big(\sigma_{11c}(\hat{q}^f_{ij})\big) + 
    \nu_c^{\geq}\big(\sigma_{11d}(\hat{q}^f_{ij})\big) \right)\\
\nu_{12}(\hat{\bm{R}}) &= 
    \frac{1}{\lvert \cC \rvert} \sum_{i \in \cC}
    \left(\nu_c^{\geq}\big(\sigma_{12a}(\hat{R}_{ij})\big) + 
    \nu_c^{\geq}\big(\sigma_{12b}(\hat{R}_{ij})\big) \right)\\
\nu_{14}(\hat{\bm{q}}^f\!, \bm{q}^f) &= 
    \frac{1}{\lvert \cP \rvert} \sum_{(i,j) \in \cP}
    \nu_c^{=}\big(\sigma_{14}(\hat{q}_{ij}^f, {q}_{ij}^f)\big)\\
\end{flalign*}
}

\section{Additional Experiments: Classification under Fairness Constraints}

This section reports additional results on LDF for constrained predictor problems on a fairness application domain.

\begin{figure}[!h]
\centering
\includegraphics[scale=.22]{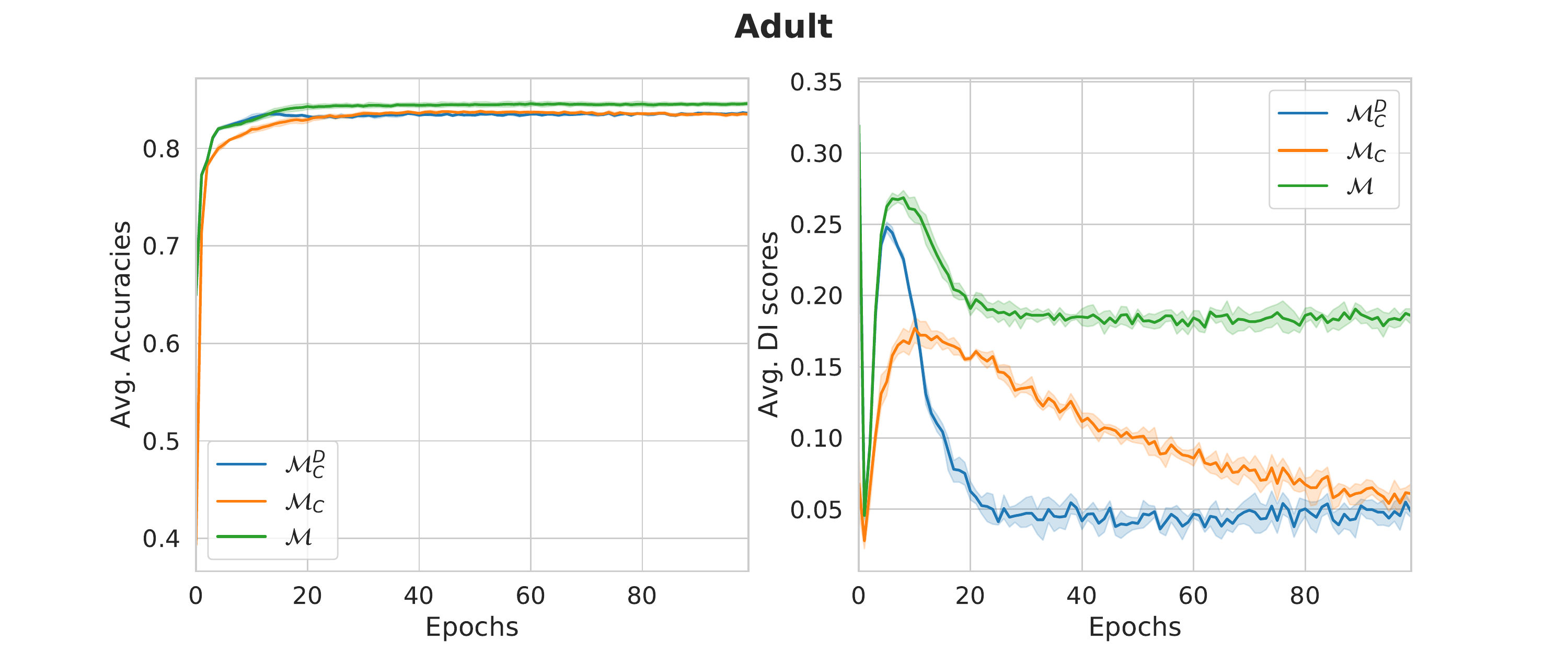}
\includegraphics[scale=.22]{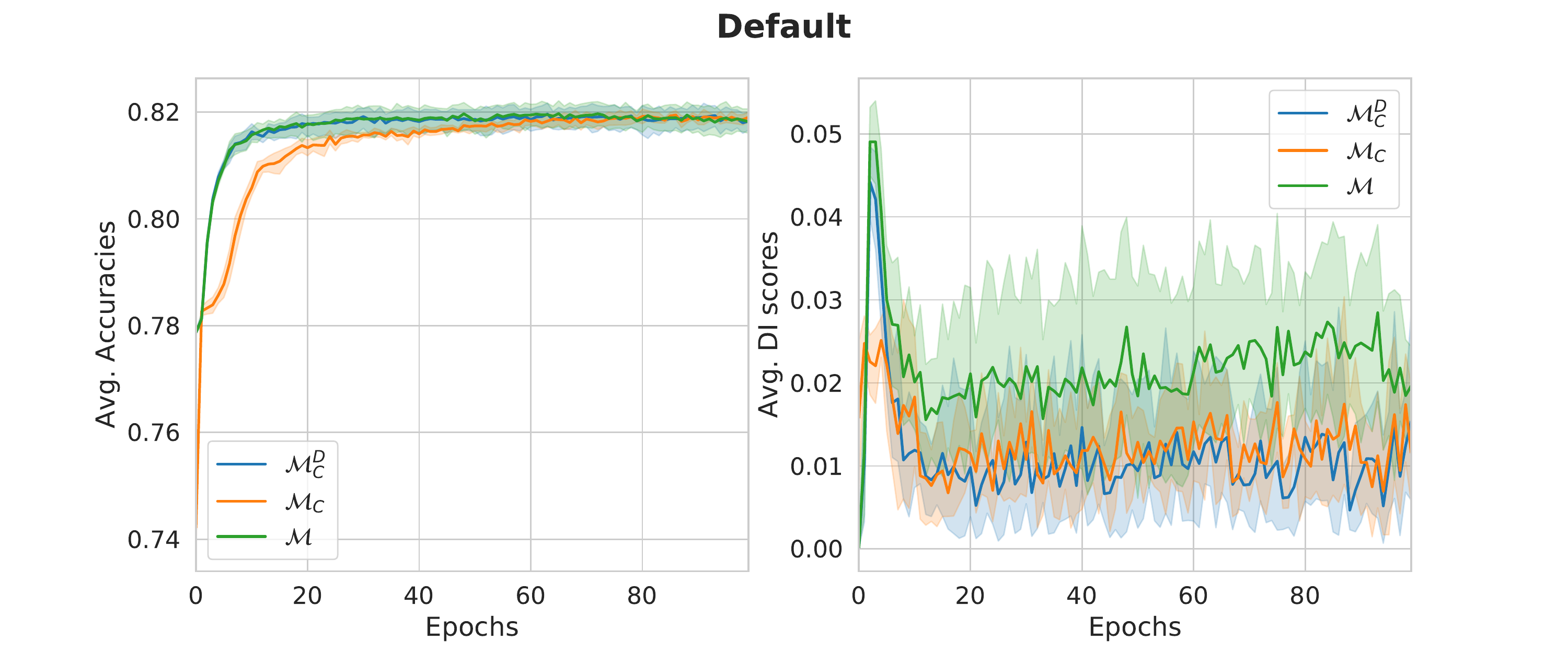}
\includegraphics[scale=.22]{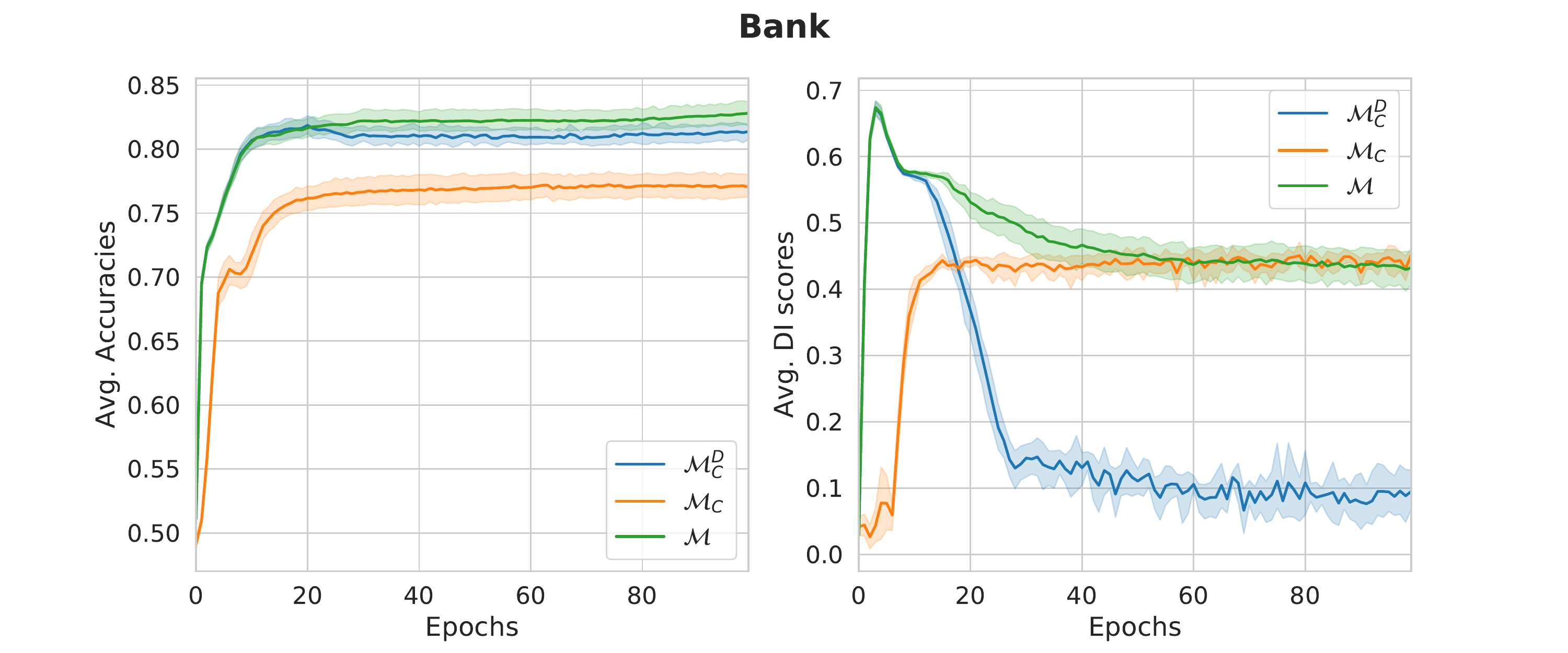}
\caption{Average (Avg.) of Accuracies/DT-scores on the validation sets during optimization.}
\label{fig:trace}
\end{figure}

Figure \ref{fig:trace} illustrates the evolution of the accuracy and {DT} metrics for all the models evaluated at the increasing of the number of epochs. The results are reported on the validation sets and the figure illustrates the results for the Adult (top) Default (middle) and Bank (bottom) datasets.
A clear trend appears: The Lagrangian Dual method $\mathcal{M}_C^D$, 
reports a lower DT scores, and this happens early in the training stages. Surprising, on the Bank dataset, the DT score could be reduced quite significantly without much loss on accuracy terms.  

\subsection{Relaxed fairness constraints}
Next, this section reports an extended evaluation on a \emph{relaxed} notion of the fairness constraints. To do so, the experiments allow the constraint a slack $0 \leq \Delta_f\leq \epsilon$ ($\epsilon \geq 0$). 
The idea is to exploit such relaxation to allow the model to achieve even better accuracy. 

The experiments report the DT and accuracy scores attained by the 
$\mathcal{M}_C^D$ model on the Adult, Default, and Bank datasets using several levels of constraint relaxation, which act on the slack $\epsilon$.   
For each dataset, the experiments compute the  
DI-score obtained in the original data, i.e $DI_{\text{org}} = |Pr(y(x) =1 |z(x) =1) -Pr(y(x) =1 |z(x) =0) $ and choose $\epsilon$ to be $5\%$, $20\%$ and $50\%$ of this quantity. 

Table \ref{tab:relaxed_fairness} reports the results. It can be observed that the DI-score increases with the increasing of the constraint relaxation $\epsilon$. At the same time, the accuracy decreases with the increasing of value $\epsilon$.

\begin{table}[!t]
\centering
\begin{adjustbox}{width=0.6\columnwidth,center}
  \begin{tabular}{l@{\hspace{6pt}} |@{\hspace{6pt}}
                 r@{\hspace{10pt}} r@{\hspace{10pt}} 
                 r@{\hspace{10pt}} r@{\hspace{10pt}} 
                 r@{\hspace{10pt}} r@{\hspace{10pt}}} 
     \toprule
     \multirow{2}{*}{\textbf{Dataset}}&  
     \multicolumn{2}{c}{$\epsilon = 5\% \text{DI}_{\text{org}}$}&
     \multicolumn{2}{c}{$\epsilon = 20\% \text{DI}_{\text{org}}$}&
     \multicolumn{2}{c}{$\epsilon = 50\% \text{DI}_{\text{org}}$}\\
      \cmidrule(r){2-3}
      \cmidrule(r){4-5}
      \cmidrule(r){6-7}
     & Acc. &  DT & Acc. &  DT & Acc. &  DT \\
     \midrule
     Adult  & 0.8336 & 0.0567 & 0.8339 &0.0586
     & 0.8358 & 0.0812\\
     Default  &0.8164 & 0.0073
                       &0.8162 & 0.0087
                       &0.8159 & 0.0092\\
     Bank   &0.8161 & 0.1556
                       &0.8215 & 0.2437
                       &0.8251& 0.3372 \\
   \bottomrule
  \end{tabular}
\end{adjustbox}
\caption{Accuracy/DT-score of $\mathcal{M}_C^D$ under different relaxed fairness parameters $\epsilon$ }
\label{tab:relaxed_fairness} 

\end{table}

%and $\bm{q}$ denoting the vector of gas (flux) flow $(q_{ij} \:|\: (i,j) \in \cC)$.

\end{document}